\pdfoutput=1

\documentclass[11pt]{article}

\usepackage{acl}

\usepackage{times}
\usepackage{latexsym}
\usepackage{mathtools}

\usepackage[T1]{fontenc}

\usepackage[utf8]{inputenc}

\usepackage{microtype}

\usepackage{graphicx}
\usepackage{amsmath}
\usepackage{multirow}
\usepackage{caption}
\usepackage{subcaption}
\usepackage{cleveref}

\newcommand{\sparkles}[0]{\includegraphics[height=.02\textwidth]{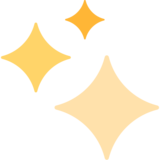}}

\newcommand{\scism}[2]{{#1}\text{e-}{#2}}

\newcommand{\Rob}{RoB}{}
\newcommand{\Roberta}{RoBERTa}{}
\newcommand{\RobB}{{\Rob}$_{\text{Base}}$}{}
\newcommand{\RobertaB}{{\Roberta}$_{\text{base}}$}{}
\newcommand{\teacher}{{\Rob}$_{\text{Large}}$}{}
\newcommand{\Teacher}{{\Roberta}$_{\text{Large}}$}{}
\newcommand{\student}{Distil{\Rob}}{}
\newcommand{\Student}{Distil{\Roberta}}{}

{}

\newcommand{\Ours}[0]{Glitter {\sparkles}}{}
\newcommand{\ours}[0]{Glitter }{}
\newcommand{\our}[0]{Glitter}{}

\title{When Chosen Wisely, More Data Is What You Need: \\A Universal Sample-Efficient Strategy For Data Augmentation}

\author{Ehsan Kamalloo\thanks{\ \ Equal Contribution.}\ \:\thanks{\ \ Work done while interning at Huawei Noah's Ark Lab.} \\
  Univeristy of Alberta \\
  \texttt{kamalloo@ualberta.ca} \\
  \And
  Mehdi Rezagholizadeh$^{*}$ \\
  Huawei Noah's Ark Lab \\
  \texttt{mehdi.rezagholizadeh@huawei.com} \\
  \AND
  Ali Ghodsi \\
  University of Waterloo \\
  \texttt{ali.ghodsi@uwaterloo.ca} \\
}

\begin{document}
\maketitle
\begin{abstract}

Data Augmentation (DA) is known to improve the generalizability of deep neural networks. Most existing DA techniques naively add a certain number of augmented samples without considering the quality and the added computational cost of these samples. To tackle this problem, a common strategy, adopted by several state-of-the-art DA methods, is to adaptively generate or re-weight augmented samples with respect to the task objective during training. However, these adaptive DA methods: (1) are computationally expensive and not sample-efficient, and (2) are designed merely for a specific setting. In this work, we present a universal DA technique, called \textit{\our}, to overcome both issues. Glitter can be plugged into any DA method,  making training sample-efficient without sacrificing performance. From a pre-generated pool of augmented samples, Glitter adaptively selects a subset of worst-case samples with maximal loss, analogous to adversarial DA. Without altering the training strategy, the task objective can be optimized on the selected subset. Our thorough experiments on the GLUE benchmark, SQuAD, and HellaSwag in three widely used training setups including consistency training, self-distillation and knowledge distillation reveal that Glitter is substantially faster to train and achieves a competitive performance, compared to strong baselines.\footnote{Our code is available at \url{https://github.com/huawei-noah/KD-NLP/tree/main/Glitter}.}

\end{abstract}

\section{Introduction}

The undeniable importance of data in deep learning~\cite{sambasivan2021everyone, rogers-2021-changing} and the costly process of data annotation has propelled researchers into leveraging Data Augmentation (DA) in a broad range of applications from computer vision~\cite{cubuk2018autoaugment, wang2020neural} to natural language processing (NLP) including machine translation~\cite{sennrich-etal-2016-improving, shen2020simple}, language understanding~\cite{shen2020simple, qu2021coda, du-etal-2021-self, kamalloo-etal-2021-far}, and question answering~\cite{alberti-etal-2019-synthetic, longpre2019exploration, shakeri-etal-2020-end}. DA is shown to be effective in improving generalization of deep neural networks~\cite{devries2017improved, xie2019unsupervised} and in increasing the number of training samples especially in low resource data regimes~\cite{sennrich-etal-2016-improving, zhang2017mixup}.
Nonetheless, in NLP, the discrete nature of text poses additional complexity to DA as generating semantically viable text from another text is challenging \cite{feng-etal-2021-survey}.

DA methods can be broadly categorized into {\em task-aware} and {\em task-agnostic} methods.
Task-agnostic DA methods essentially generate augmented text regardless of the task at hand and often do not warrant additional training or fine-tuning. They can be based on some hand-crafted heuristics \cite{zhang2015character, wei-zou-2019-eda}, back-translation \cite{sennrich-etal-2016-improving, edunov-etal-2018-understanding}, or token replacement from a pre-trained language model \cite{kobayashi-2018-contextual, wu2019conditional, ng-etal-2020-ssmba}.
Even though deploying task-agnostic methods is straightforward, these methods do not take into account any task-specific information, and thus, their performance is usually limited. On the other hand, task-aware DA methods are capable of generating augmented samples, conditioned on the downstream task objective~\cite{hu2019learning, xie2019unsupervised, rashid-etal-2021-mate}. These methods adapt augmented examples specifically for a task in that they construct augmented examples, sometimes partly, during training.
Despite their advantages, they often incur additional training costs, resulting in a prohibitively slow and a computationally expensive training.

In general, the central problems surrounding DA techniques in NLP can be summarized as follows:
First, DA methods are mostly not sample-efficient in that they add arbitrary number of augmented samples to the training data %
and naively incorporate all of them into training without investigating how many of augmented samples are actually needed. Second, although more effective, task-aware methods are notoriously time-consuming to train. This is especially problematic in large-scale datasets such as SQuAD~\cite{rajpurkar-etal-2016-squad} and MNLI~\cite{williams-etal-2018-broad}.
Third, most DA methods are not universal as they work solely with a particular setup---e.g., training a single-network~\cite{xie2019unsupervised}, or training in teacher-student settings~\cite{rashid-etal-2021-mate}. Overall, the importance of both sample efficiency and training efficiency for DA has been often overlooked.

Motivated by the above problems, in this work, we introduce a universal DA method, {\Ours}\footnote{Inspired by ``{\em All that is gold does not glitter}'' ---J.R.R. Tolkien, The Fellowship of the Ring.}, which can be plugged into any DA method to make them sample-efficient, and task-aware without sacrificing performance.
Specifically, given a pool of augmented samples that are generated offline, our proposed method follows a minimax approach \cite{farnia2016minimax} to select a small subset with maximal expected loss ({\em maximization step}) during training. Without any further adjustments to the training algorithm, the task objective can be optimized for this selected subset ({\em minimization step}).

Our key contributions in this paper can be summarized as follows: 
\begin{enumerate}
    \item {\ours} is a universal method which can be effortlessly applied to any DA method to enforce sample efficiency while maintaining (or even boosting) their performance. 
    \item We devise strategies to adapt {\ours} for a variety of widely used training setups including single-network, consistency training, self-distillation and knowledge distillation. 
    \item Through our empirical evaluations, we show that Glitter achieves superior performance over state-of-the-art DA methods on GLUE, SQuAD, and HellaSwag, while significantly speeding up the training.
\end{enumerate}

\section{Related Work}

\subsection{Task-agnostic DA in NLP}

Contextual augmentation techniques~\cite{kobayashi-2018-contextual, wu2019conditional} use pre-trained language models for DA. \citet{kobayashi-2018-contextual} propose bidirectional LSTM language models for word substitution conditioned on the label of their input text.   
SSMBA~\cite{ng-etal-2020-ssmba} and TinyBERT~\cite{jiao-etal-2020-tinybert} perturb the input by masking some of the tokens, and then, sample tokens from a BERT model to replace the masked tokens and generate augmented samples.   
Back-Translation~\cite{sennrich-etal-2016-improving} augments data using two consecutive translation models: the first model to translate the input  into an arbitrary target language; then, a second model to translate the result back into its original language.  
Mixed-up~\cite{guo2019augmenting} generates augmented samples based on interpolating word embedding and sentence embedding vectors.
\citet{shen2020simple} introduce a set of cut-off techniques that zero out contiguous spans of the embedding matrix at token level, feature level and span level.
EDA~\cite{wei-zou-2019-eda} consists of simple word-level operations including synonym replacement, random deleting, random insertion and random swapping.
\subsection{Task-aware DA in NLP}

One approach to leverage task-specific information is to assign different weights to augmented samples based on their individual impacts on the model \cite{yi2021reweighting}.
Although effective, the re-weighting mechanism largely ignores sample efficiency.  
\citet{wu2019conditional} introduce a mask-and-reconstruct approach, namely c-BERT, that fine-tune a pre-trained BERT model to predict label-compatible tokens. 
CoDA \cite{qu2021coda} combines various label-preserving transformations with adversarial training jointly with a contrastive regularization objective.
Unsupervised DA (UDA; \citealt{xie2019unsupervised}) uses off-the-shelf DA methods and adds an auxiliary \textit{consistency loss} to the training objective. However, UDA is not sample-efficient and it is designed only for a single-network setup; how to deploy it in other training scenarios such as knowledge distillation is not clear. 
\citet{hu2019learning} propose a reinforcement learning-based technique where the reward function is defined based on whether generated augmented samples are label-preserving or not.

\subsection{DA for KD}
KD~\cite{bucilua2006model,kd}, initially proposed as a model compression technique, aims at transferring the knowledge of an already trained model, called {\em teacher}, to a smaller or a same-size {\em student} model. 
Several studies found that DA can significantly boost KD's performance in NLP. TinyBERT~\cite{jiao-etal-2020-tinybert} uses a task-agnostic DA technique for its task-specific fine-tuning. \citet{kamalloo-etal-2021-far} and \citet{rashid-etal-2021-mate} showed that DA can also be tailored for KD. In particular, MATE-KD \cite{rashid-etal-2021-mate} tunes a separate masked language model in order to  generate augmented samples with maximum divergence. \citet{kamalloo-etal-2021-far} and \citet{du-etal-2021-self} employ $k$NN retrieval to fetch augmented samples from a massive sentence bank.

Glitter differs from previous work in that it simultaneously focuses on sample efficiency, and universality such that it can be freely used in any training setting.

\begin{figure*}[h]
    \centering
    \includegraphics[width=0.8\textwidth]{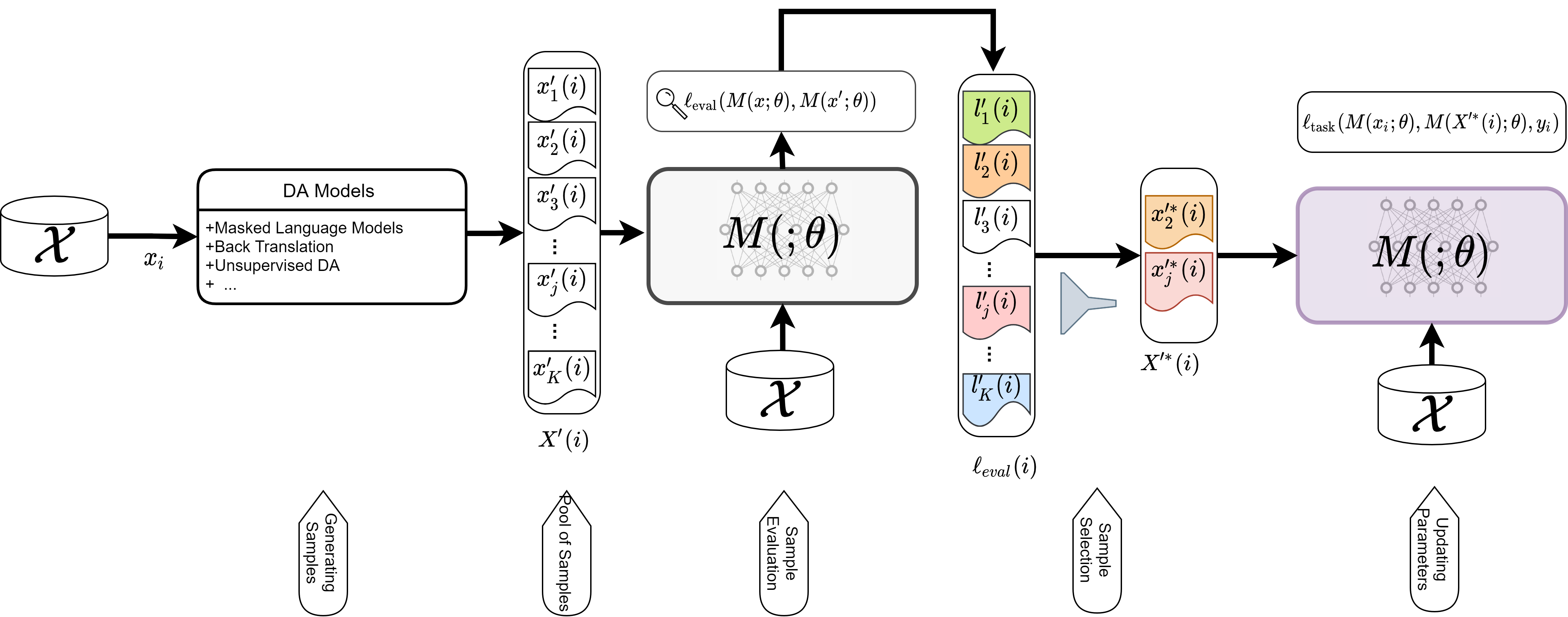}
    \caption{Illustration of {\Ours} (from left to right): first, generating augmented samples from different DA techniques; second, forming a pool of samples $X'(i)$; third, evaluating the augmented samples using the $\ell_{eval}()$ loss; fourth, filtering the top-$k_1$ samples based on their corresponding $\ell_{eval}()$; fifth, updating the parameters of the model by minimizing the task loss $\ell_\text{task}(:\theta)$.     } 
    \label{fig:fig1}
\end{figure*}

\section{Methodology}

In this section, we introduce our task-aware DA method, {\Ours}, that aims at using an efficient number of augmented samples without sacrificing performance.
Our proposed strategy is agnostic to DA methods; it can be seamlessly plugged into any DA method with any training setting to enforce sample efficiency.

Existing learning-based DA methods train a separate DA model and adapt its output for a particular objective function that is entirely task-dependent: 
\begin{equation}
    \begin{split}
    & \phi^* \leftarrow \underset{\phi}{\min} ~ \ell_{DA}(M(\Omega(x;\phi);\theta)) \\
    &    x'^* = \Omega(x;\phi^*)
    \end{split}
    \label{eq:DA:task-aware}
\end{equation}
where $\ell_{DA}()$ is a loss function, geared towards the objective of the task, $\Omega(;\phi)$ is the DA model with trainable parameters $\phi$, and $M(;\theta)$ refers to the original model, parameterized by $\theta$.

In contrast to learning-based DA, we propose to generate many augmented candidates using any arbitrary DA method prior training, and adaptively select most suitable candidates during training. This procedure does not introduce additional trainable parameters into training, and more importantly, is capable of automatically ignoring unnecessary augmented examples.
Let $(x_i,y_i)_{i=1}^{N} \in \{(\mathcal{X},\mathcal{Y})\}$ represent training data such that a pair $x_i \in \mathcal{X}$ and $y_i \in \mathcal{Y}$ are an input example and its corresponding label.
Suppose a pool of $K$ augmented examples, $X'(i)=\{x'_k(i)\}_{k=1}^K$, are sampled from some DA model for each training example $(x_i,y_i) \in (\mathcal{X},\mathcal{Y})$.
Note that {\our} imposes no restrictions on how to augment training data; augmented samples can be generated via a single or even multiple DA models. 
\paragraph{Sample Selection.} Given a pool of augmented samples, our approach is to adaptively select the best candidates according to particular defined criteria. Inspired by the minimax approach~\cite{farnia2016minimax, volpi2018generalizing}, our selection mechanism is based on finding top-$k_1$ (out of $K$) worst-case augmented samples from the $X'$ set. Minimizing the main model loss function on these worst-case augmented samples will help improving generalization of the model~\cite{volpi2018generalizing}.
In order to rank augmented samples, we evaluate $X'(i)$ based on a distance function with respect to the corresponding original training sample, $x_i$, within the model's latent space:

\begin{equation}
    \label{eq:DA:task-aware}
    \begin{split}
    & X'^*(i) \leftarrow \text{top}_{k_1}\Big(\ell_{\text{eval}} \big(M(x_i;\theta),M(X'(i);\theta)\big) \Big) \\ 
    &  X'^*(i) = \{ x'^*_j(i) \}_{j=1}^{k_1} \subset X'(i)
    \end{split}
\end{equation}
where $\text{top}_{k_1}()$ denotes returns top-$k_1$ indices based on the scores returned by $\ell_{\text{eval}}$, 
$X'^*(i)$ is the set of $k_1$ selected augmented samples for $x_i$; $\ell_{eval}()$ is the evaluation loss which is determined via the task objective.
\paragraph{Updating the Model Parameters.} After obtaining the top-$k_1$ augmented samples, we group them with the original training samples, $\{x_i\} \cup X'^*(i)$, and subsequently, update the model parameters only based on this selected set of augmented samples on the original loss: 
\begin{equation}
    \begin{split}
    \mathcal{L}(\theta) &= \sum_{i=1}^{N} \ell_{\text{task}}\Big(M(x_i;\theta), M(X'^*(i);\theta), y_i\Big)\\
    \theta_{t} &\leftarrow \theta_{t-1} - \lambda \nabla_\theta (\mathcal{L}(\theta))|_{\theta_{t-1}}
    \end{split}
\end{equation}
where $N$ is the number of training samples, $\lambda$ is the learning rate, and $\ell_{\text{task}}()$ is the final task loss---e.g., cross entropy (ce) for classification---that is computed over both original data and selected augmented data.
In the remainder of this section, we discuss how {\ours} can be applied to popular training settings including general DA for single networks, and DA for teacher-student (KD) setups. Note that {\ours} is not restricted to these settings and may be adapted for other settings such as DAIR \cite{huang2022dair}.

\subsection{General DA for Single Networks} 
We consider three potential setups for the single network scenario: (1) General single network, (2) Self-distillation, and (3) Consistency training.
\paragraph{General Single Network.}
In this setup, augmented samples are exploited in a semi-supervised manner where we can evaluate them based on the divergence of their predicted output $M(x'_k(i);\theta)=p(y|x'_k(i);\theta)$ from the ground-truth label or the prediction of the original corresponding training sample $M(x_i;\theta)=p(y|x_i;\theta)$ using the cross entropy loss, $\ell_{ce}$: 
\begin{equation}
    \begin{split}
    &    \ell_{\text{eval}} = \ell_{ce}\big(y_i, M(x'_k(i);\theta)\big) \\ 
    & \text{or} \\ 
    &   \ell_{\text{eval}} = \ell_{ce}\big(M(x_i;\theta), M(x'_k(i);\theta)\big). 
    \end{split}
    \label{eq:general:eval}
\end{equation}

The cross entropy criterion is not the only option here. Other choices for $\ell_{\text{eval}}$ include (but not limited to) focal loss \cite{lin2017focal}, and tilted loss \cite{li2021tilted}.

For the final task loss, $\ell_{\text{task}}$ we can deploy a standard cross entropy loss over both training samples and their corresponding selected augmented samples:  
\begin{equation}
    \begin{split}
    &    \ell_{\text{task}} = \ell_{ce}\big(y_i, M(x_i;\theta)\big) +\\
    &  \frac{1}{k_1} \sum_{x\in X'^*(i)} \ell_{ce}\big(y_i, M(x;\theta)\big). 
    \end{split}
\end{equation}

\paragraph{Consistency Training (CT; \citealt{xie2019unsupervised}).} In this configuration, we can employ the same $\ell_{\text{eval}}$ introduced in Eq.~\eqref{eq:general:eval}. As a result, our method naturally selects top-$k_1$ most inconsistent augmented samples for each training sample.
Then, the network is optimized to make predictions for input augmented samples that are consistent with predictions of their corresponding original training samples:
\begin{equation}
    \begin{split}
    &    \ell_{\text{task}}^{\text{CT}} =  \ell_{ce}\big(y_i, M(x_i;\theta_t)\big) + \\ 
    & \frac{1}{k_1}\sum_{x\in  X'^*(i)} \ell_{ce}\big( M(x_i;\theta_{t-1}),M(x;\theta_{t})\big). 
    \end{split}
    \label{eq:ct:task}
\end{equation}
As stated by~\citet{xie2019unsupervised}, the second term in Eq.~\eqref{eq:ct:task} leverages the previous prediction of the network for each training example.

\paragraph{Self-Distillation (Self-KD).} In Self-KD, we first train a model, and then, use it ($M(;\theta^*)$) as a teacher to train an identical model but initialized from scratch using KD \cite{furlanello2018born}. How to adjust $\ell_{\text{eval}}$ and $\ell_{\text{task}}$ is detailed in \S\ref{KD}.

\subsection{DA for Teacher-Student (KD)}
\label{KD}
In this setup, we have a teacher model, $T(;\psi^*)$ with parameters $\psi$ that is already trained on the training data, along with a student model, $M(;\theta)$, which we aim to train. The selection criterion for augmented samples is to maximize divergence between the teacher and the student:  
\begin{equation}
    \begin{split}
    &    \ell_{\text{eval}}^{\text{KD}} = \ell_{KL}\Big(T\big(x'_k(i);\psi^*\big), M\big(x'_k(i);\theta\big)\Big) \\ 
    \end{split}
    \label{eq:KD:eval}
\end{equation}
where $\ell_{KL}$ refers to the KL divergence. After selecting the maximum divergence augmented samples, then we calculate the KD loss as following: 
\begin{equation}
    \begin{split}
    &    \ell_{\text{task}}^{\text{KD}} = \alpha~ \ell_{ce}\big(y_i, M(x_i;\theta)\big) + (1-\alpha) \times\\ 
    & \frac{1}{k_1+1}\sum_{x\in \{x_i\} \cup X'^*(i)} \ell_{KL}\big( T(x;\psi^*),M(x;\theta)\big) 
    \end{split}
    \label{eq:KD:task}
\end{equation}
where $\alpha$ is a hyperparameter. 

\section{Experiments} \label{sec:experiments}

\subsection{Setup}
To incorporate unlabelled augmented data into training, we adopt CT \cite{xie2019unsupervised} and KD \cite{kd}.
To this end, we conduct experiments under two settings:

\paragraph{Standalone} where we train a single model on the augmented data. In this setting, we seek to answer two questions: (1) How much is DA capable of improving the model generalization? (2) Does sample efficiency of Glitter hurt performance? For this purpose, we fine-tune {\RobertaB} \cite{liu2019roberta} using CT and Self-KD on augmented data.

\paragraph{Distilled} where we distill {\Student} \cite{sanh2019distilbert} (student) from {\Teacher} \cite{liu2019roberta} (teacher) using the augmented data. Note that the teacher is already trained on the original data and DA comes into play only during distilling the student model. Our goal here is to investigate whether DA is an effective means in knowledge transfer to curb the capacity gap \cite{cho2019efficacy} between a large model and a small one.

In both settings, we take the best performing model on the development set and evaluate it on the test set (depicted by \textit{Test}). Additionally, for the standalone model setting, we also report results on the development set when models are trained only for 5 epochs (depicted by \textit{Dev}), similar to CoDA \cite{qu2021coda}, to make a comparison with baselines. Our \textit{Dev} results are an average of 10 runs with different seeds.
The implementation details and hyperparameters are provided in~\S\ref{sec:supp_implement}.

\subsubsection{DA Methods}
We leverage three widely used textual augmentation methods:
\begin{enumerate}
    \item \textbf{EDA} \cite{wei-zou-2019-eda}\footnote{\url{https://github.com/makcedward/nlpaug}}:
    We randomly replace 5\% of the tokens with their synonyms and randomly delete up to 10\%.
    \item {\bf Back-Translation (BT; \citealt{sennrich-etal-2016-improving})}: We use fairseq \cite{ott-etal-2019-fairseq} to translate sentences into German and then back into English. We do nucleus sampling \cite{holtzman2019curious} with $p = 0.9$ for both translations. We find that $p = 0.6$ works better on sentiment classification.
    \item {\bf Mask-and-Reconstruct (MR; \citealt{ng-etal-2020-ssmba})}: We randomly mask 15\% of the tokens and construct a new sentence by sampling from a pre-trained BERT$_\text{Large}$ for masked tokens. We adopt top-$k$ sampling with $k=20$ to select new tokens. For MNLI, we obtain better results with top-10 sampling.
\end{enumerate}
For each augmentation method, we generate 12 augmented examples per training instance for all datasets, except for large datasets---i.e., MNLI, QQP, and SQuAD---where the number of augmented examples are 8 per train example.

\subsubsection{Baselines}
Because the two environments---i.e., standalone and distilled---are different in nature, we compare {\our} with different baselines for each environment. For both, Vanilla-DA that takes all augmented data into account without reservation is the first baseline.

The baselines for the standalone setting are: CoDA \cite{qu2021coda}, MMEL \cite{yi2021reweighting}, and HiddenCut \cite{chen-etal-2021-hiddencut}.
And for distilled, we consider MATE-KD \cite{rashid-etal-2021-mate}.

\begin{table*}[t]
    \centering
    \small
    \begin{tabular}{l c c c c c c c c c}
        \hline
        \multirow{2}{*}{{\bf Method}} & \small{{\bf CoLA}} & \small{{\bf SST}} & \small{{\bf MRPC}} & \small{{\bf STS-B}} & \small{{\bf QQP}} & \small{{\bf MNLI-m/mm}} & \small{{\bf QNLI}} & \small{{\bf RTE}} & \multirow{2}{*}{{\bf Avg.}} \\
        & \small{Mcc} & \small{Acc} & \small{Acc/F$_1$} & \small{P/S} & \small{Acc/F$_1$} & \small{Acc} & \small{Acc} & \small{Acc} & \\
        \hline
        {\teacher} (teacher) & 63.8 & 96.8 & 90.6 & 92.4 & 81.5 & 90.3/89.8 & 94.8 & 88.3 & 87.3 \\
        BERT$_\text{Large}$ $^{\scriptscriptstyle \clubsuit}$ & 60.5 & 94.9 & 87.4 & 87.1 & 80.7 & 86.7/85.9 & 92.7 & 70.1 & 82.5  \\
        \hline
        {\student} & 55.2 & 93.9 & 85.9 & 86.0 & 80.3 & 84.0/83.1 & 90.6 & 73.6 & 81.1 \\
        KD & 54.9 & 94.0 & 86.8 & 87.3 & 80.5 & 85.1/83.7 & 91.9 & 73.5 & 81.7 \\
        \hline
        \multicolumn{10}{c}{{\em Task-Aware DA}} \\
        MATE-KD $^{\scriptscriptstyle \clubsuit}$ & 56.0 & 94.9 & {\bf 90.2} & {\bf 88.0} & {\bf 81.2} & 85.5/84.8 & 92.1 & {\bf 75.0} & \underline{82.8} \\
        \hline
        \multicolumn{10}{c}{{\em EDA} \cite{wei-zou-2019-eda}} \\
        Vanilla-DA (8x) & 55.5 & 94.8 & 87.6 & 86.1 & 80.7 & 85.3/84.7 & 92.0 & 72.8 & 81.8 \\
        {\Ours} & 54.5 & {\bf 95.1} & 87.5 & 86.5 & 80.4 & 85.4/84.8 & 92.1 & 73.2 & 81.8 \\
        & \small{{\em 8x/2x}} & \small{{\em 8x/1x}} & \small{{\em 8x/2x}} & \small{{\em 8x/2x}} & \small{{\em 8x/2x}} & \small{{\em 8x/2x}} & \small{{\em 8x/2x}} & \small{{\em 8x/1x}} & \\
        \hline
        \multicolumn{10}{c}{{\em Back-Translation}} \\
        Vanilla-DA (8x) & 53.4 & {\bf 95.1} & 88.5 & 87.5 & 80.9 & 85.9/{\bf 85.9} & \underline{92.2} & 73.5 & 82.1 \\
        {\Ours} & 54.9 & {\bf 95.1} & 88.4 & 87.3 & 80.9 & \underline{86.2}/\underline{85.3} & \underline{92.2} & 73.7 & 82.3 \\
        & \small{{\em 8x/2x}} & \small{{\em 8x/1x}} & \small{{\em 8x/1x}} & \small{{\em 8x/2x}} & \small{{\em 8x/2x}} & \small{{\em 8x/2x}} & \small{{\em 8x/2x}} & \small{{\em 8x/2x}} & \\
        \hline
        \multicolumn{10}{c}{{\em Mask-and-reconstruct}} \\
        Vanilla-DA (8x) & \underline{58.8} & 94.5 & 88.7 & 87.0 & 80.9 & 85.8/84.9 & 91.8 & 74.0 & 82.6 \\
        {\Ours} & \textbf{59.2} & {\bf 95.1} & \underline{89.2} & \underline{87.6} & \underline{81.0} & {\bf 86.6}/84.8 & {\bf 92.4} & \underline{74.1} & {\bf 83.0}  \\
        & \small{{\em 8x/1x}} & \small{{\em 8x/1x}} & \small{{\em 8x/2x}} & \small{{\em 8x/1x}} & \small{{\em 8x/2x}} & \small{{\em 8x/2x}} & \small{{\em 8x/2x}} & \small{{\em 8x/2x}} & \\
        \hline
    \end{tabular}
    \caption{Test results of the distilled experiment on GLUE. ($^{\scriptscriptstyle \clubsuit}$) denotes results are taken verbatim from: BERT$_\text{Large}$ \cite{devlin-etal-2019-bert}, and MATE-KD \cite{rashid-etal-2021-mate}. {\bf Bold} and \underline{underlined} numbers indicate the best and the second best results across the DA methods.}
    \label{tab:glue_test}
\end{table*}

\begin{table*}[t]
    \centering
    \small
    \begin{tabular}{l c c c c c c c c c}
        \hline
        \multirow{2}{*}{\small{{\bf Method}}} & \small{{\bf CoLA}} & \small{{\bf SST}} & \small{{\bf MRPC}} & \small{{\bf STS-B}} & \small{{\bf QQP}} & \small{{\bf MNLI-m}} & \small{{\bf QNLI}} & \small{{\bf RTE}} & \multirow{2}{*}{{\bf Avg.}} \\
         & \small{Mcc} & \small{Acc} & \small{Acc/F$_1$} & \small{P/S} & \small{Acc/F$_1$} & \small{Acc} & \small{Acc} & \small{Acc} & \\
        \hline
        {\small {\Roberta}} & 61.9 & 95.4 & 88.6 & 89.3 & 80.4 & 87.6 & 93.0 & 81.6 & 84.7 \\
        \hline
        {\small Self-KD} & 61.7 & 95.7 & 89.0 & 89.0 & 80.8 & {\bf 88.3} & 93.0 & 81.7 & 84.9 \\
        {\small \, + Vanilla-DA} & 61.5 & {\bf 96.1} & 88.9 & {\bf 89.7} & 81.0 & 88.0 & 92.9 & 81.1 & 84.9 \\
        & \small{{\em 8x}} & \small{{\em 8x}} & \small{{\em 8x}} & \small{{\em 8x}} & \small{{\em 8x}} & \small{{\em 8x}} & \small{{\em 8x}} & \small{{\em 12x}} & \\
        {\small \, + {\Ours}} & 62.5 & 96.0 & {\bf 89.8} & 89.5 & {\bf 81.1} & 88.1 & {\bf 93.5} & {\bf 82.3} & {\bf 85.4} \\
        & \small{{\em 8x/1x}} & \small{{\em 8x/2x}} & \small{{\em 8x/2x}} & \small{{\em 8x/2x}} & \small{{\em 8x/2x}} & \small{{\em 8x/2x}} & \small{{\em 8x/2x}} & \small{{\em 12x/1x}} & \\
        \hline
        {\small CT + Vanilla-DA} & 59.4 & 95.6 & 89.0 & 85.8 & 80.3 & 82.5 & 92.0 & 80.2 & 83.1 \\
        & \small{{\em 8x}} & \small{{\em 8x}} & \small{{\em 8x}} & \small{{\em 10x}} & \small{{\em 8x}} & \small{{\em 8x}} & \small{{\em 8x}} & \small{{\em 10x}} & \\
        {\small CT + {\Ours}} & {\bf 62.7} & 95.8 & 89.2 & 87.9 & 80.9 & 84.1 & 92.9 & 81.8 & 84.4 \\
        & \small{{\em 8x/1x}} & \small{{\em 8x/1x}} & \small{{\em 8x/1x}} & \small{{\em 10x/1x}} & \small{{\em 8x/2x}} & \small{{\em 8x/2x}} & \small{{\em 8x/2x}} & \small{{\em 10x/1x}} & \\
        \hline
    \end{tabular}
    \caption{Test result of the standalone experiments on GLUE using {\RobertaB}.}
    \label{tab:glue_test_base}
\end{table*}

\begin{table*}[t]
    \small
    \centering
    \begin{tabular}{l c c c c c | c c c}
        \hline
        \multirow{2}{*}{\small{{\bf Method}}} & \footnotesize{{\bf SST}} & \small{{\bf MRPC}} & \small{{\bf MNLI-m}} & \small{{\bf QNLI}} & \footnotesize{{\bf RTE}} & \small{{\bf IMDb-Con.}} & \small{{\bf A-NLI}} & \small{{\bf HANS}} \\
          & \small{Acc} & \small{F$_1$} & \small{Acc} & \small{Acc} & \small{Acc} & \small{Acc} & \small{Acc} & \small{Acc} \\
        \hline
        {\small {\Rob}}$^{\scriptscriptstyle \spadesuit}$ & 94.8 & 90.2 & 87.6 & 92.8 & 78.7 & - & - & - \\
        {\small CoDA}$^{\scriptscriptstyle \spadesuit}$ & 95.3 & \small{91.7} & \small{88.1} & 93.6 & 82.0 & - & - & - \\
        {\small HiddenCut}$^{\scriptscriptstyle \spadesuit}$ & 95.8 & \small{92.0} & {\bf \small{88.2}} & {\bf 93.7} & 83.4 & 87.8 & {\bf 32.8} & 71.2  \\
        {\small MMEL}$^{\scriptscriptstyle \dagger}$ & 94.6 {\scriptsize $\pm$ 0.8} & 91.9 {\scriptsize $\pm$ 0.4} & 88.1 {\scriptsize $\pm$ 0.1} & 93.2 {\scriptsize $\pm$ 0.1} & 85.3 {\scriptsize $\pm$ 1.0} & 90.5 {\scriptsize $\pm$ 0.7} & 31.4 {\scriptsize $\pm$ 0.6} & 74.5 {\scriptsize $\pm$ 0.6} \\
        \hline
        {\small {\Rob}}$^{\scriptscriptstyle \dagger}$ & 94.3 {\scriptsize $\pm$ 0.1} & 91.6 {\scriptsize $\pm$ 0.5} & 87.7 {\scriptsize $\pm$ 0.1} & 92.8 {\scriptsize $\pm$ 0.2} & 84.5 {\scriptsize $\pm$ 0.8} & 90.0 {\scriptsize $\pm$ 0.4} & 30.8 {\scriptsize $\pm$ 0.9} & 73.6 {\scriptsize $\pm$ 0.7} \\
        {\small Self-KD} & 94.3 {\scriptsize $\pm$ 0.2} & 91.5 {\scriptsize $\pm$ 0.3} & 87.9 {\scriptsize $\pm$ 0.1} & 92.9 {\scriptsize $\pm$ 0.2} & 84.0 {\scriptsize $\pm$ 0.6} & 90.3 {\scriptsize $\pm$ 0.5} & 30.9 {\scriptsize $\pm$ 0.4} & 73.5 {\scriptsize $\pm$ 0.7} \\
        {\; + Vanilla-DA} & 95.4 {\scriptsize $\pm$ 0.5} & 92.0 {\scriptsize $\pm$ 0.3} & {\bf 88.2 {\scriptsize $\pm$ 0.1}} & 93.4 {\scriptsize $\pm$ 0.1} & 84.4 {\scriptsize $\pm$ 0.7} & 90.2 {\scriptsize $\pm$ 0.4} & 31.3 {\scriptsize $\pm$ 0.5} & 73.9 {\scriptsize $\pm$ 0.4} \\
        {\; + {\Ours}} & {\bf 95.7 {\scriptsize $\pm$ 0.2}} & {\bf 92.2 {\scriptsize $\pm$ 0.5}} & {\bf 88.2 {\scriptsize $\pm$ 0.1}} & 93.4 {\scriptsize $\pm$ 0.1} & {\bf 85.6 {\scriptsize $\pm$ 0.7}} & {\bf 90.6 {\scriptsize $\pm$ 0.2}} & 31.8 {\scriptsize $\pm$ 0.4} & {\bf 74.6 {\scriptsize $\pm$ 0.3}}  \\
        \hline
    \end{tabular}
    \caption{Dev results of the standalone experiment on GLUE using {\RobertaB}. ($^{\scriptscriptstyle \spadesuit}$) denotes results are taken verbatim from: {\Rob} and CoDA \cite{qu2021coda}, and HiddenCut \cite{chen-etal-2021-hiddencut}. ($^{\scriptscriptstyle \dagger}$) indicates the results are obtained from our implementation of MMEL \cite{yi2021reweighting}.}
    \label{tab:glue_dev_base}
\end{table*}

\subsection{GLUE} \label{sec:glue}
The GLUE benchmark \cite{wang2018glue} is a well-known suite of nine\footnote{We excluded WNLI since our DA methods are not designed for this task.} tasks that aim at evaluating natural language understanding models. We present test results in the distilled mode in \Cref{tab:glue_test}.
{\our} consistently outperforms Vanilla-DA, while it is faster to train. Specifically, {\our} achieves parity with Vanilla-DA for EDA in terms of the overall average score, while scoring +0.2\% and +0.4\% higher for BT and MR, respectively. We observe that only in few cases Vanilla-DA negligibly outperforms {\our}---e.g., on MRPC, and STS-B for BT. Nonetheless, {\our} \emph{8x/1x} trains 50\% faster than Vanilla-DA \emph{8x} on average, and 30\% faster for \emph{8x/2x}. 
Also, {\our} surpasses MATE-KD by +0.2\% in the overall score. Unlike {\our}, MATE-KD introduces additional parameters to the model during training and it trains drastically slower because it generates augmented examples on-the-fly.
Moreover, \Cref{tab:glue_test} illustrates that MR yields the best test results across the three DA methods except for SST where BT leads to better results. Based on this observation, we report results on MR augmented data for all GLUE datasets except for SST in the remainder of our experiments.

For the standalone mode, \Cref{tab:glue_test_base,tab:glue_dev_base} present the results on test and dev, respectively. Similar to distilled, {\our} outperforms Vanilla-DA by +0.5\% for both self-KD and CT. Self-KD yields better results than CT on all GLUE tasks except CoLA. CT falls short on most GLUE tasks, compared to no DA results---i.e., top-2 rows in \Cref{tab:glue_test_base}. This is why, we only evaluated {\our} with self-KD on the dev data. {\our} achieves superior performance gains, compared to all three baselines on all datasets except QNLI. The key advantage of {\our} is that the training procedure remains intact.

\subsubsection{Out-of-Domain Generalization}
\label{ssec:ood_glue}
We also evaluate {\our} on OOD datasets. To this end, we test our models, already trained on GLUE tasks, on OOD datasets whose data distribution differs from the original data. In particular, here are our selected OOD datasets:
\begin{itemize}
    \item SST: IMDb \cite{maas-etal-2011-learning}, IMDb-Cont. \cite{gardner-etal-2020-evaluating}, and IMDb-CAD \cite{Kaushik2020Learning}, as done in \citet{chen-etal-2021-hiddencut}. Although both SST and IMDb datasets are collected on movie reviews, IMDb reviews tend to be substantially longer than SST sentences.
    \item STS-B: SICK \cite{marelli-etal-2014-sick}, a semantic relatedness dataset, created from image and video captions. SICK and STS-B are collected on roughly identical domains, but from different sources.
    \item QQP: PAWS$_\text{QQP}$ \cite{zhang-etal-2019-paws}, analogous to \citet{chen-etal-2021-hiddencut}, and MQP \cite{mccreery2020effective}, a medical question similarity dataset.
    \item MNLI: SciTail \cite{khot2018scitail}, collected from school-level science questions, and similar to \citet{chen-etal-2021-hiddencut}, A-NLI \cite{nie-etal-2020-adversarial}, and HANS \cite{mccoy-etal-2019-right}.
    \item RTE: HANS \cite{mccoy-etal-2019-right}.
\end{itemize}

\Cref{tab:glue_ood_kd} in \S\ref{ssec:ood_distilled} showcases the OOD results for the distilled mode. {\our} outperforms Vanilla-DA in most cases, and is on par with it for nearly the rest. The only exceptions are IMDb-Cont., MQP, and PAWS$_\text{QQP}$ where Vanilla-DA outperforms {\our} by almost 1\% on average. Also, all models do not generalize well to PAWS$_\text{QQP}$ and A-NLI because their performance is below a majority-class performance. Moreover, a fine-tuned {\Student} achieves the best OOD performance on HANS, highlighting that DA is not actually helpful for OOD accuracy on HANS.

\Cref{tab:glue_dev_base} (the right side) reports the OOD results for standalone models. The complete results are presented in \S\ref{ssec:ood_standalone}---i.e., \Cref{tab:glue_test_ood_self} on test and \Cref{tab:glue_dev_ood_self} on dev. {\our} overwhelmingly outperforms all the baselines with a few exceptions. In the dev results, the fine-tuned model with no DA achieves the best OOD generalization on IMDb, and SciTail, while HiddenCut scores the highest on A-NLI with a 1\% margin. Similarly, in the test results, {\our} trails Self-KD with no DA on IMDb, IMDb-CAD, and SciTail.

\begin{table}[t]
    \centering
    \begin{tabular}{l c c}
        \hline
        \multirow{2}{*}{{\bf Method}} & {\bf SQuAD} & {\bf HellaSwag} \\
        & \small{EM/F$_1$} & \small{Acc} \\
        \hline
        {\teacher} & 88.9/94.6 & 85.2 \\
        \hline
        {\student} & 80.9/87.9 & 42.9 \\
        KD & 81.1/88.2 & 42.5 \\
        \, + Vanilla-DA {\em \scriptsize (8x)} & 81.8/89.1 & 41.8 \\
        \, + {\Ours} {\em \scriptsize (8x/2x)} & {\bf 83.6/90.3} & {\bf 44.1} \\
        \hline
    \end{tabular}
    \caption{Dev results of the distilled experiment on two downstream tasks.}
    \label{tab:other_id_results}
\end{table}

\subsection{HellaSwag}
HellaSwag \cite{zellers-etal-2019-hellaswag} is a dataset for situated commonsense reasoning that involves picking the best ending given a context. We augment contexts in HellaSwag using only BT to ensure that the choices remain meaningful for the augmented contexts. Because our standalone results have been consistent with the distilled results, we report our results only in the distilled mode.
According to our results demonstrated in \Cref{tab:other_id_results}, {\our} comfortably surpasses Vanilla-DA by a +2.3\% margin.

\subsection{SQuAD}
SQuAD \cite{rajpurkar-etal-2016-squad} is a crowd-sourced reading comprehension benchmark that consists of more than 100K questions, derived from Wikipedia passages. The task objective is to extract an answer span from a given question/passage pair. We augment questions in SQuAD v1.1 using only BT to ensure that the answer can still be found in the given passage for the augmented questions. Analogous to HellaSwag, we report our results only in the distilled mode. As shown in \Cref{tab:other_id_results}, {\our} outperformas Vanilla-DA by +1.8\% in exact-match accuracy on the development set.

We also evaluate our trained models under distribution shift by testing them on QA datasets from four different domains: Wikipedia, New York Times, Reddit, and Amazon product reviews \cite{miller2020effect}. The OOD results are presented in \Cref{tab:squad_ood_results}. {\our} is consistently superior to Vanilla-DA in all four domains.

\begin{table}[t]
    \small
    \centering
    \begin{tabular}{l c c c c}
        \hline
        \multirow{2}{*}{{\bf Method}} & \small{{\bf Wiki}} & \small{{\bf NYT}} & \small{{\bf Reddit}} & \small{\textbf{Amzn}} \\
        & \scriptsize{EM} & \scriptsize{EM} & \scriptsize{EM} & \scriptsize{EM} \\
        \hline
        {\teacher} & 84.4 & 85.9 & 76.6 & 74.4 \\
        \hline
        {\student} & 76.6 & 78.1 & 66.2 & 62.9 \\
        KD & 76.5 & 78.7 & 65.7 & 63.0 \\
        \, + Vanilla-DA & 77.3 & 79.0 & 65.9 & 63.3 \\
        \, + {\Ours} & {\bf 79.3} & {\bf 80.7} & {\bf 68.1} & {\bf 64.7} \\
        \hline
    \end{tabular}
    \caption{OOD results for models trained on SQuAD and tested on QA datasets from four different domains \cite{miller2020effect}.}
    \label{tab:squad_ood_results}
\end{table}

\section{Ablation Study and Discussion}
\label{sec:discussion}
In this section, we aim to answer the following questions:

\begin{itemize}
    \item How does training time of {\our} compare against Vanilla-DA?
    \item Instead of adaptively selecting augmented data during training, can we pre-process them to dispense with unnecessary examples prior to training?
    \item How many augmented examples are required for {\our} to work?
    \item Is our selection strategy based on sorting of $\ell_{eval}$ in {\our} important?
\end{itemize}

For this purpose, we conduct a detailed analysis on 4 GLUE tasks---i.e., SST, MRPC, QNLI, and RTE. We trained models based on Vanilla-DA and {\our} using Self-KD and tested them on the development set (the dev setting).

\begin{figure*}[t]
     \centering
     \begin{subfigure}[b]{0.24\textwidth}
         \centering
         \includegraphics[width=\textwidth]{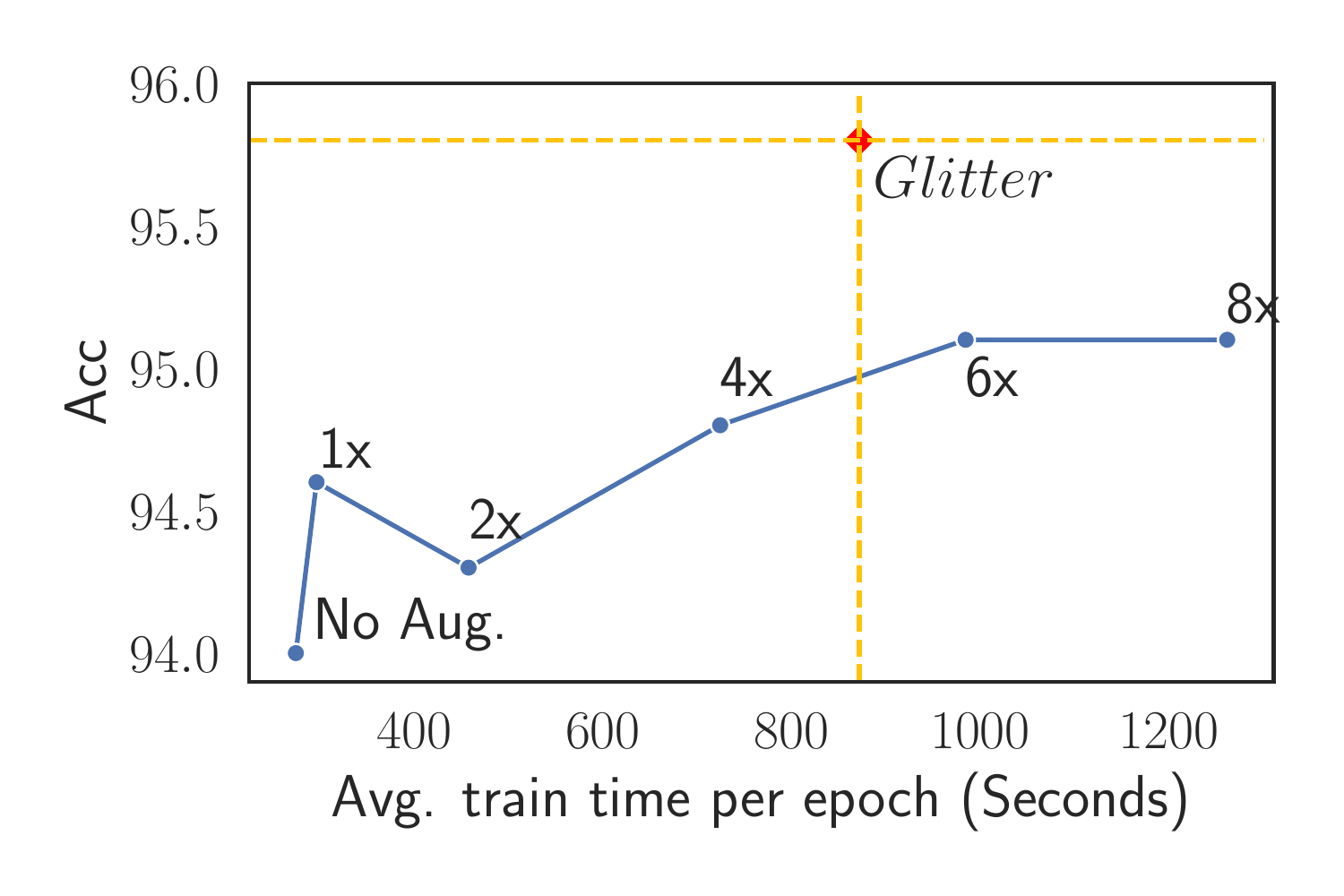}
         \caption{SST}
         \label{fig:SST_runtime}
     \end{subfigure}
     \hfill
     \begin{subfigure}[b]{0.24\textwidth}
         \centering
         \includegraphics[width=\textwidth]{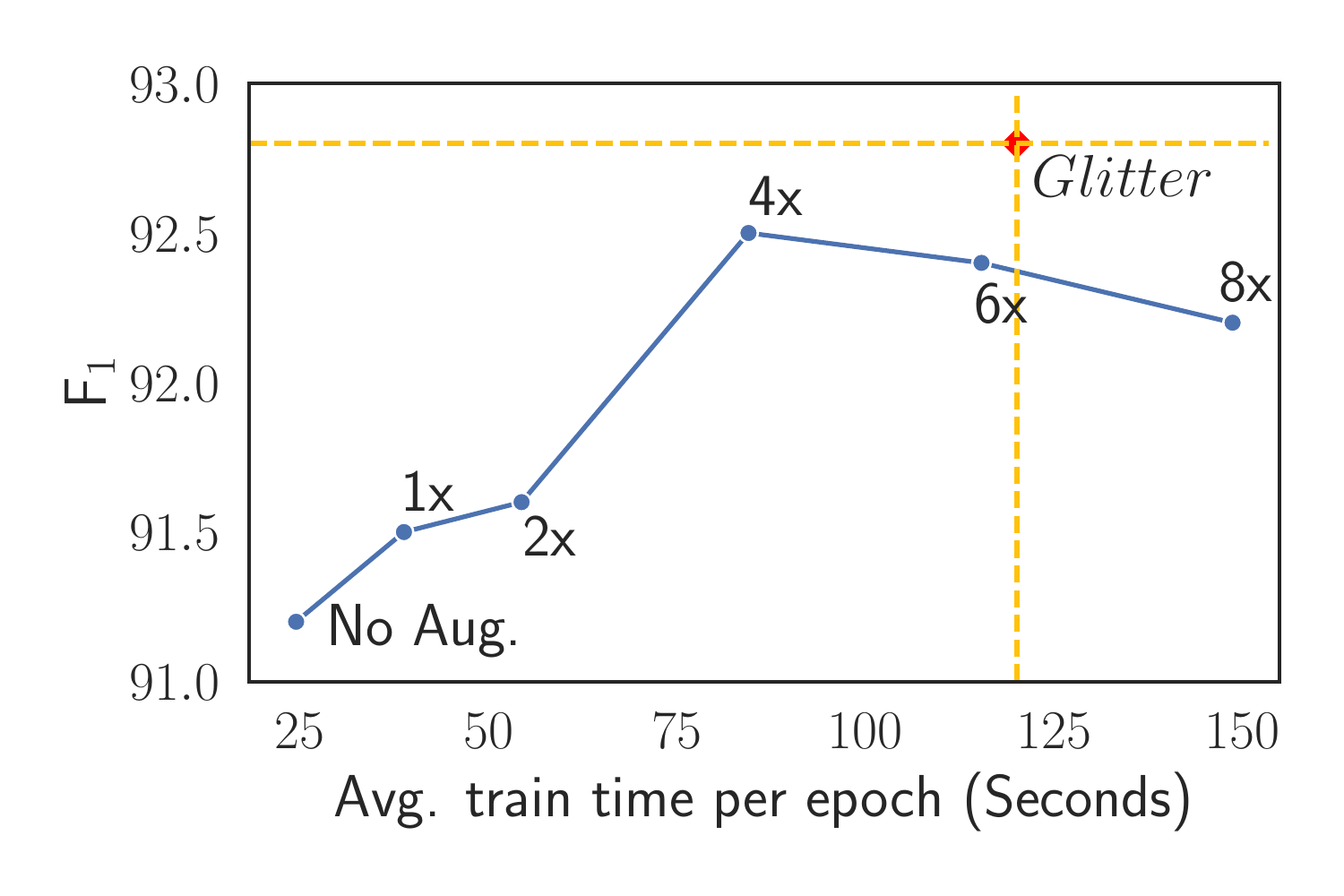}
         \caption{MRPC}
         \label{fig:MRPC_runtime}
     \end{subfigure}
     \hfill
     \begin{subfigure}[b]{0.24\textwidth}
        \centering
        \includegraphics[width=\textwidth]{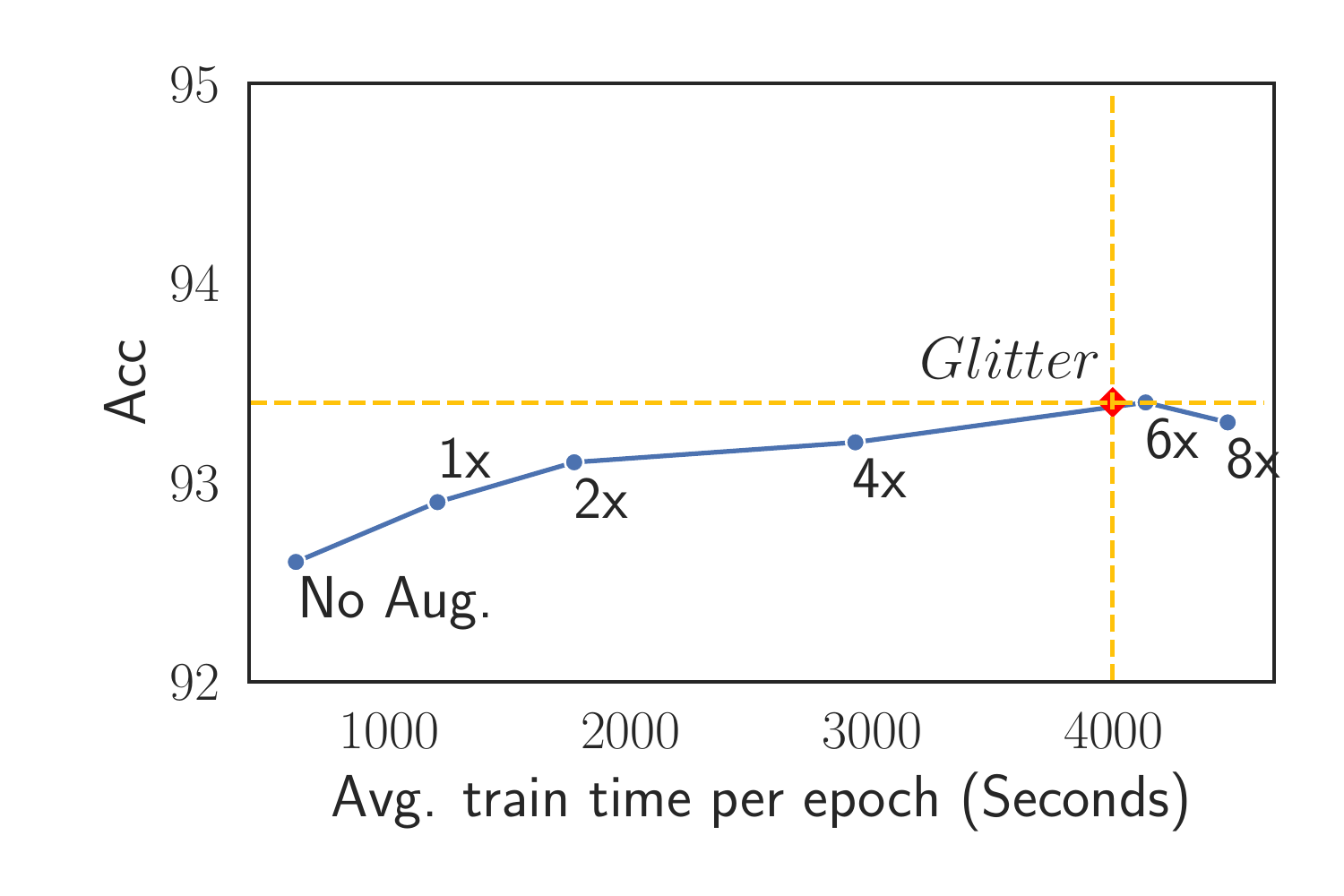}
        \caption{QNLI}
        \label{fig:QNLI_runtime}
    \end{subfigure}
     \hfill
     \begin{subfigure}[b]{0.24\textwidth}
         \centering
         \includegraphics[width=\textwidth]{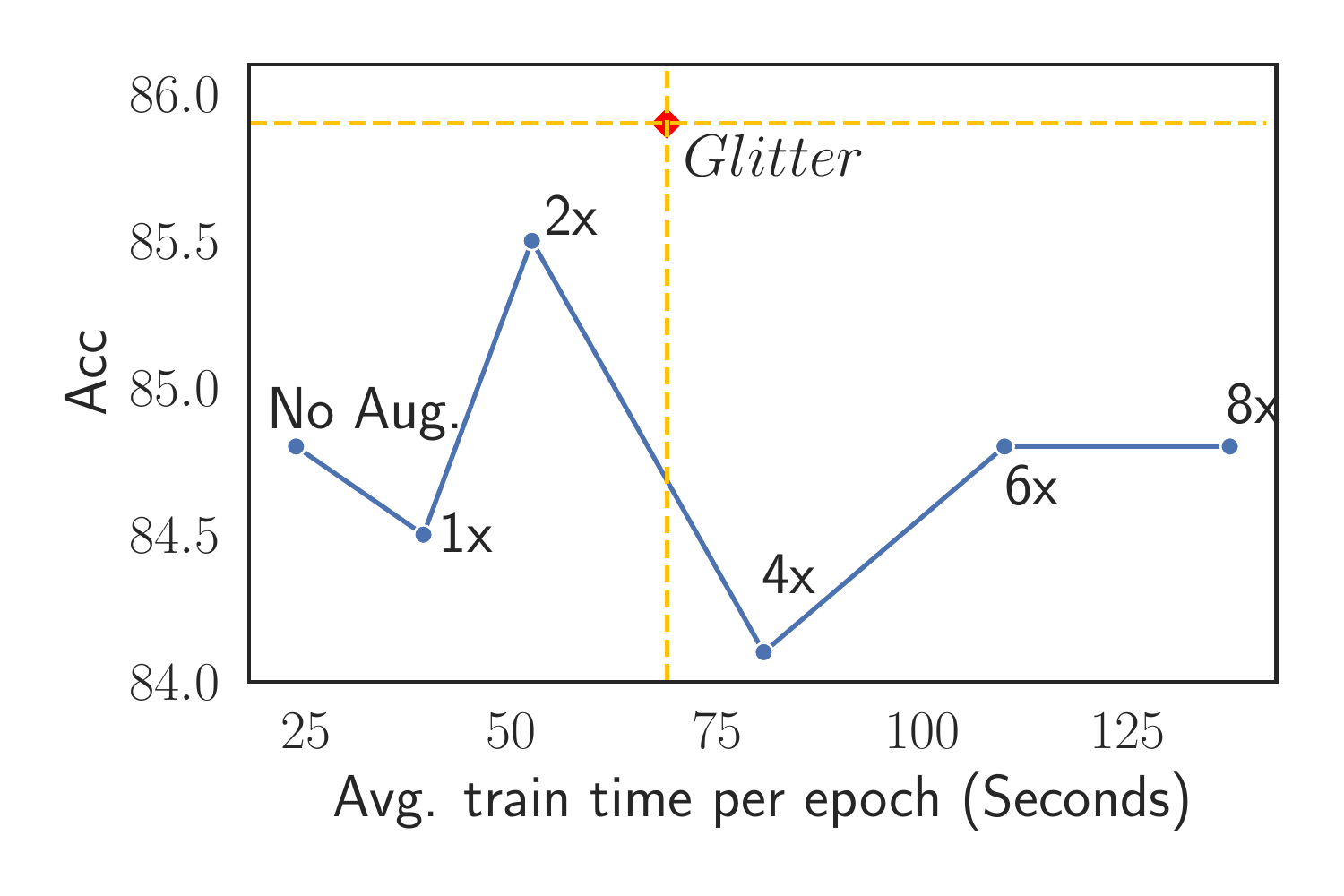}
         \caption{RTE}
         \label{fig:RTE_runtime}
     \end{subfigure}
        \caption{Runtime Analysis of DA when training {\RobertaB} using self-KD. The red point signifies {\our}.}
        \label{fig:runtime}
\end{figure*}

\paragraph{Runtime Analysis.}
Throughout our experiments in \S\ref{sec:experiments}, we compare Glitter with Vanilla-DA when number of augmentations are similar for both methods---i.e., {\em 8x}. A natural question is: how would both DA methods behave with fewer augmented data? To this end, we vary augmentation size from $1x$ to $8x$ and train different Vanilla-DA models on each augmented dataset. We measure average the training time per epoch for all models. \Cref{fig:runtime} illustrates the dev accuracy as the training time increases. The training speed of {\our} {\em 8x/2x} is slightly faster than Vanilla-DA {\em 6x} on SST, MRPC, and QNLI and for {\our} {\em 8x/1x}, is faster than Vanilla-DA {\em 4x} on RTE. {\our} is superior of the two on all datasets.

\paragraph{Effect of Pre-processing Augmented Data.}
We conjecture that {\our} does not need any data engineering on augmented examples to obtain preferable performance gains. However, Vanilla-DA may require some pre-processing by weeding out potentially noisy data to become more effective. To investigate this, we exploit two pre-processing techniques: {\bf (1) Confidence-based filtering:} Augmented examples for which the model's confidence is below a minimum threshold $\beta$ are discarded, {\bf (2) Label-preserving augmentation (LP):} Augmented examples for which the model predicts a different label than the original example are discarded. 
The results, reported in \Cref{tab:ablation_filter_strategies}, show no meaningful performance gains by these pre-processing techniques. For Vanilla-DA, minimum confidence threshold of $0.7$ performs slightly better as it brings minor improvements on MRPC (+0.3\%) and QNLI (+0.1\%), but is still lower than {\our}. On the other hand, applying these techniques slightly deteriorates the performance of {\our} in almost all cases. The only improvements are +0.1\% on QNLI for LP and $\beta$=0.7.

\begin{table}[t]
    \small
    \centering
    \begin{tabular}{l c c c c}
        \hline
        \multirow{2}{*}{{\bf Method}} & \small{{\bf SST}} & \small{{\bf MRPC}} & \small{{\bf QNLI}} & \small{\textbf{RTE}} \\
        & \scriptsize{Acc} & \scriptsize{F$_1$} & \scriptsize{Acc} & \scriptsize{Acc} \\
        \hline
        Vanilla-DA & 95.1 & 92.2 & 93.3 & 84.8 \\
        \,\, $\beta = 0.7$ & 95.1 & 92.5 & 93.4 & 84.8 \\
        \,\, $\beta = 0.9$ & 95.0 & 92.2 & 93.3 & 83.8 \\
        \,\, LP & 94.8 & 92.4 & 93.3 & 84.8 \\
        \hline
        {\Ours} & 95.8 & 92.8 & 93.4 & 85.9 \\
        \,\, $\beta = 0.7$ & 95.0 & 91.5 & 93.5 & 85.2 \\
        \,\, $\beta = 0.9$ & 95.0 & 92.5 & 93.3 & 84.1 \\
        \,\, LP & {95.1} & 92.2 & 93.5 & {85.9} \\
        \hline
    \end{tabular}
    \caption{Dev results of self-KD exhibiting the effectiveness of different pre-processing techniques to filter augmented examples on 4 GLUE tasks. $\beta$ and LP depict a minimum confidence threshold, and label preserving, respectively.}
    \label{tab:ablation_filter_strategies}
\end{table}

\paragraph{Effect of Augmentation Size in {\our}.}
We explore how augmentation size affects the performance of {\our}. Throughout our experiments, we fix the augmentation size to {\em 8x}, but now, we reduce augmentation size $K$ to $6x$ and $4x$, while retaining selection size $k_1$ as before---i.e., 1 for RTE, and 2 for the rest. Our results, shown in \Cref{tab:ablation_glitter}, reveal that when $K$ becomes close to $k_1$, {\our}'s performance declines. Nonetheless, for a sufficiently large augmentation, {\our} starts to shine. For SST, and MRPC, the magic number is $8x$, whereas for QNLI, and RTE, {\our} performs best on $6x$. Another parameter in Glitter is the selection size $k_1$. We find that for all tasks, the best value can be chosen from $\{1, 2\}$ (2 by default). Using this method, tuning $k_1$ is straightforward and does not impose additional complexity to our method.

\paragraph{Effect of Selection Strategy in {\our}.}
In this section, our objective is to assess whether our proposed selection algorithm is crucial in {\our}. To this end, we sample random augmented examples at each iteration, namely {\em Glitter-Rnd}, instead of selecting worst-case examples. As illustrated in \Cref{tab:ablation_glitter} (the bottom two rows), the performance drops on all datasets---i.e., 0.2\% on QNLI, and more than 1\% on the rest, confirming the effectiveness of our selection algorithm.

\begin{table}[t]
    \small
    \centering
    \begin{tabular}{l c c c c}
        \hline
        \multirow{2}{*}{{\bf Method}} & \small{{\bf SST}} & \small{{\bf MRPC}} & \small{{\bf QNLI}} & \small{\textbf{RTE}} \\
        & \scriptsize{Acc} & \scriptsize{F$_1$} & \scriptsize{Acc} & \scriptsize{Acc} \\
        \hline
        {\Ours} (8x) & 95.8 & 92.8 & 93.4 & 85.9 \\
        {\Ours} (6x) & 94.7 & 92.7 & 93.7 & 86.3 \\
        {\Ours} (4x) & 95.0 & 92.1 & 93.3 & 85.7 \\
        Glitter-Rnd (8x/2x) & 94.3 & 91.4 & 93.2 & 85.2 \\
        Glitter-Rnd (8x/1x) & 94.3 & 91.8 & 93.2 & 84.5 \\
        \hline
    \end{tabular}
    \caption{Dev results of self-KD for studying the effect of augmentation size and the selection algorithm for 4 GLUE tasks.}
    \label{tab:ablation_glitter}
\end{table}

\section{Conclusion}
In this work, 
we proposed a universal DA technique, namely \textit{\our}, that can be freely applied to any DA technique to enforce sample efficiency without introducing additional parameters or changing the training procedure. 
We extensively evaluated {\ours} on a broad range of NLU tasks and in various widely used settings including consistency training, self-distillation and knowledge distillation and demonstrated substantial efficiency gains without compromising effectiveness.
Extending Glitter to auto-regressive models for machine translation and abstractive summarization is an interesting direction for future work.

\bibliography{anthology,main}

\begin{thebibliography}{58}
\expandafter\ifx\csname natexlab\endcsname\relax\def\natexlab#1{#1}\fi

\bibitem[{Alberti et~al.(2019)Alberti, Andor, Pitler, Devlin, and
  Collins}]{alberti-etal-2019-synthetic}
Chris Alberti, Daniel Andor, Emily Pitler, Jacob Devlin, and Michael Collins.
  2019.
\newblock \href {https://doi.org/10.18653/v1/P19-1620} {Synthetic {QA} corpora
  generation with roundtrip consistency}.
\newblock In \emph{Proceedings of the 57th Annual Meeting of the Association
  for Computational Linguistics}, pages 6168--6173, Florence, Italy.
  Association for Computational Linguistics.

\bibitem[{Buciluǎ et~al.(2006)Buciluǎ, Caruana, and
  Niculescu-Mizil}]{bucilua2006model}
Cristian Buciluǎ, Rich Caruana, and Alexandru Niculescu-Mizil. 2006.
\newblock Model compression.
\newblock In \emph{Proceedings of the 12th ACM SIGKDD international conference
  on Knowledge discovery and data mining}, pages 535--541.

\bibitem[{Chen et~al.(2021)Chen, Shen, Chen, and
  Yang}]{chen-etal-2021-hiddencut}
Jiaao Chen, Dinghan Shen, Weizhu Chen, and Diyi Yang. 2021.
\newblock \href {https://doi.org/10.18653/v1/2021.acl-long.338} {{H}idden{C}ut:
  Simple data augmentation for natural language understanding with better
  generalizability}.
\newblock In \emph{Proceedings of the 59th Annual Meeting of the Association
  for Computational Linguistics and the 11th International Joint Conference on
  Natural Language Processing (Volume 1: Long Papers)}, pages 4380--4390,
  Online. Association for Computational Linguistics.

\bibitem[{Cho and Hariharan(2019)}]{cho2019efficacy}
Jang~Hyun Cho and Bharath Hariharan. 2019.
\newblock On the efficacy of knowledge distillation.
\newblock In \emph{Proceedings of the IEEE/CVF International Conference on
  Computer Vision (ICCV)}, pages 4794--4802.

\bibitem[{Cubuk et~al.(2019)Cubuk, Zoph, Mane, Vasudevan, and
  Le}]{cubuk2018autoaugment}
Ekin~D Cubuk, Barret Zoph, Dandelion Mane, Vijay Vasudevan, and Quoc~V Le.
  2019.
\newblock {AutoAugment}: Learning augmentation policies from data.
\newblock In \emph{Proceedings of the IEEE/CVF Conference on Computer Vision
  and Pattern Recognition (CVPR)}.

\bibitem[{Devlin et~al.(2019)Devlin, Chang, Lee, and
  Toutanova}]{devlin-etal-2019-bert}
Jacob Devlin, Ming-Wei Chang, Kenton Lee, and Kristina Toutanova. 2019.
\newblock \href {https://doi.org/10.18653/v1/N19-1423} {{BERT}: Pre-training of
  deep bidirectional transformers for language understanding}.
\newblock In \emph{Proceedings of the 2019 Conference of the North {A}merican
  Chapter of the Association for Computational Linguistics: Human Language
  Technologies, Volume 1 (Long and Short Papers)}, pages 4171--4186,
  Minneapolis, Minnesota. Association for Computational Linguistics.

\bibitem[{DeVries and Taylor(2017)}]{devries2017improved}
Terrance DeVries and Graham~W Taylor. 2017.
\newblock Improved regularization of convolutional neural networks with cutout.
\newblock \emph{arXiv preprint arXiv:1708.04552}.

\bibitem[{Du et~al.(2021)Du, Grave, Gunel, Chaudhary, Celebi, Auli, Stoyanov,
  and Conneau}]{du-etal-2021-self}
Jingfei Du, Edouard Grave, Beliz Gunel, Vishrav Chaudhary, Onur Celebi, Michael
  Auli, Veselin Stoyanov, and Alexis Conneau. 2021.
\newblock \href {https://doi.org/10.18653/v1/2021.naacl-main.426}
  {Self-training improves pre-training for natural language understanding}.
\newblock In \emph{Proceedings of the 2021 Conference of the North American
  Chapter of the Association for Computational Linguistics: Human Language
  Technologies}, pages 5408--5418, Online. Association for Computational
  Linguistics.

\bibitem[{Edunov et~al.(2018)Edunov, Ott, Auli, and
  Grangier}]{edunov-etal-2018-understanding}
Sergey Edunov, Myle Ott, Michael Auli, and David Grangier. 2018.
\newblock \href {https://doi.org/10.18653/v1/D18-1045} {Understanding
  back-translation at scale}.
\newblock In \emph{Proceedings of the 2018 Conference on Empirical Methods in
  Natural Language Processing}, pages 489--500, Brussels, Belgium. Association
  for Computational Linguistics.

\bibitem[{Farnia and Tse(2016)}]{farnia2016minimax}
Farzan Farnia and David Tse. 2016.
\newblock A minimax approach to supervised learning.
\newblock \emph{Advances in Neural Information Processing Systems},
  29:4240--4248.

\bibitem[{Feng et~al.(2021)Feng, Gangal, Wei, Chandar, Vosoughi, Mitamura, and
  Hovy}]{feng-etal-2021-survey}
Steven Feng, Varun Gangal, Jason Wei, Sarath Chandar, Soroush Vosoughi, Teruko
  Mitamura, and Eduard Hovy. 2021.
\newblock \href {https://doi.org/10.18653/v1/2021.findings-acl.84} {A survey of
  data augmentation approaches for {NLP}}.
\newblock In \emph{Findings of the Association for Computational Linguistics:
  ACL-IJCNLP 2021}, pages 968--988, Online. Association for Computational
  Linguistics.

\bibitem[{Furlanello et~al.(2018)Furlanello, Lipton, Tschannen, Itti, and
  Anandkumar}]{furlanello2018born}
Tommaso Furlanello, Zachary Lipton, Michael Tschannen, Laurent Itti, and Anima
  Anandkumar. 2018.
\newblock \href {https://proceedings.mlr.press/v80/furlanello18a.html} {Born
  again neural networks}.
\newblock In \emph{Proceedings of the 35th International Conference on Machine
  Learning}, pages 1607--1616. PMLR.

\bibitem[{Gardner et~al.(2020)Gardner, Artzi, Basmov, Berant, Bogin, Chen,
  Dasigi, Dua, Elazar, Gottumukkala, Gupta, Hajishirzi, Ilharco, Khashabi, Lin,
  Liu, Liu, Mulcaire, Ning, Singh, Smith, Subramanian, Tsarfaty, Wallace,
  Zhang, and Zhou}]{gardner-etal-2020-evaluating}
Matt Gardner, Yoav Artzi, Victoria Basmov, Jonathan Berant, Ben Bogin, Sihao
  Chen, Pradeep Dasigi, Dheeru Dua, Yanai Elazar, Ananth Gottumukkala, Nitish
  Gupta, Hannaneh Hajishirzi, Gabriel Ilharco, Daniel Khashabi, Kevin Lin,
  Jiangming Liu, Nelson~F. Liu, Phoebe Mulcaire, Qiang Ning, Sameer Singh,
  Noah~A. Smith, Sanjay Subramanian, Reut Tsarfaty, Eric Wallace, Ally Zhang,
  and Ben Zhou. 2020.
\newblock \href {https://doi.org/10.18653/v1/2020.findings-emnlp.117}
  {Evaluating models{'} local decision boundaries via contrast sets}.
\newblock In \emph{Findings of the Association for Computational Linguistics:
  EMNLP 2020}, pages 1307--1323, Online. Association for Computational
  Linguistics.

\bibitem[{Guo et~al.(2019)Guo, Mao, and Zhang}]{guo2019augmenting}
Hongyu Guo, Yongyi Mao, and Richong Zhang. 2019.
\newblock Augmenting data with mixup for sentence classification: An empirical
  study.
\newblock \emph{arXiv preprint arXiv:1905.08941}.

\bibitem[{Hinton et~al.(2015)Hinton, Vinyals, and Dean}]{kd}
Geoffrey Hinton, Oriol Vinyals, and Jeff Dean. 2015.
\newblock Distilling the knowledge in a neural network.
\newblock \emph{arXiv preprint arXiv:1503.02531}.

\bibitem[{Holtzman et~al.(2020)Holtzman, Buys, Du, Forbes, and
  Choi}]{holtzman2019curious}
Ari Holtzman, Jan Buys, Li~Du, Maxwell Forbes, and Yejin Choi. 2020.
\newblock \href {https://openreview.net/forum?id=rygGQyrFvH} {The curious case
  of neural text degeneration}.
\newblock In \emph{International Conference on Learning Representations}.

\bibitem[{Hu et~al.(2019)Hu, Tan, Salakhutdinov, Mitchell, and
  Xing}]{hu2019learning}
Zhiting Hu, Bowen Tan, Ruslan Salakhutdinov, Tom Mitchell, and Eric~P Xing.
  2019.
\newblock Learning data manipulation for augmentation and weighting.
\newblock \emph{arXiv preprint arXiv:1910.12795}.

\bibitem[{Huang et~al.(2022)Huang, Halbe, Sankar, Amini, Kottur, Geramifard,
  Razaviyayn, and Beirami}]{huang2022dair}
Tianjian Huang, Shaunak Halbe, Chinnadhurai Sankar, Pooyan Amini, Satwik
  Kottur, Alborz Geramifard, Meisam Razaviyayn, and Ahmad Beirami. 2022.
\newblock \href {https://openreview.net/forum?id=PKdNRKjwL4} {{DAIR}: Data
  augmented invariant regularization}.
\newblock In \emph{International Conference on Learning Representations}.

\bibitem[{Jiao et~al.(2020)Jiao, Yin, Shang, Jiang, Chen, Li, Wang, and
  Liu}]{jiao-etal-2020-tinybert}
Xiaoqi Jiao, Yichun Yin, Lifeng Shang, Xin Jiang, Xiao Chen, Linlin Li, Fang
  Wang, and Qun Liu. 2020.
\newblock \href {https://doi.org/10.18653/v1/2020.findings-emnlp.372}
  {{T}iny{BERT}: Distilling {BERT} for natural language understanding}.
\newblock In \emph{Findings of the Association for Computational Linguistics:
  EMNLP 2020}, pages 4163--4174, Online. Association for Computational
  Linguistics.

\bibitem[{Kamalloo et~al.(2021)Kamalloo, Rezagholizadeh, Passban, and
  Ghodsi}]{kamalloo-etal-2021-far}
Ehsan Kamalloo, Mehdi Rezagholizadeh, Peyman Passban, and Ali Ghodsi. 2021.
\newblock \href {https://doi.org/10.18653/v1/2021.findings-acl.309} {Not far
  away, not so close: Sample efficient nearest neighbour data augmentation via
  {M}ini{M}ax}.
\newblock In \emph{Findings of the Association for Computational Linguistics:
  ACL-IJCNLP 2021}, pages 3522--3533, Online. Association for Computational
  Linguistics.

\bibitem[{Kaushik et~al.(2020)Kaushik, Hovy, and Lipton}]{Kaushik2020Learning}
Divyansh Kaushik, Eduard Hovy, and Zachary Lipton. 2020.
\newblock \href {https://openreview.net/forum?id=Sklgs0NFvr} {Learning the
  difference that makes a difference with counterfactually-augmented data}.
\newblock In \emph{International Conference on Learning Representations}.

\bibitem[{Khot et~al.(2018)Khot, Sabharwal, and Clark}]{khot2018scitail}
Tushar Khot, Ashish Sabharwal, and Peter Clark. 2018.
\newblock {SciTail}: A textual entailment dataset from science question
  answering.
\newblock In \emph{Thirty-Second AAAI Conference on Artificial Intelligence}.

\bibitem[{Kobayashi(2018)}]{kobayashi-2018-contextual}
Sosuke Kobayashi. 2018.
\newblock \href {https://doi.org/10.18653/v1/N18-2072} {Contextual
  augmentation: Data augmentation by words with paradigmatic relations}.
\newblock In \emph{Proceedings of the 2018 Conference of the North {A}merican
  Chapter of the Association for Computational Linguistics: Human Language
  Technologies, Volume 2 (Short Papers)}, pages 452--457, New Orleans,
  Louisiana. Association for Computational Linguistics.

\bibitem[{Li et~al.(2021)Li, Beirami, Sanjabi, and Smith}]{li2021tilted}
Tian Li, Ahmad Beirami, Maziar Sanjabi, and Virginia Smith. 2021.
\newblock \href {https://openreview.net/forum?id=K5YasWXZT3O} {Tilted empirical
  risk minimization}.
\newblock In \emph{International Conference on Learning Representations}.

\bibitem[{Lin et~al.(2017)Lin, Goyal, Girshick, He, and
  Doll{\'a}r}]{lin2017focal}
Tsung-Yi Lin, Priya Goyal, Ross Girshick, Kaiming He, and Piotr Doll{\'a}r.
  2017.
\newblock Focal loss for dense object detection.
\newblock In \emph{Proceedings of the IEEE international conference on computer
  vision}, pages 2980--2988.

\bibitem[{Liu et~al.(2019)Liu, Ott, Goyal, Du, Joshi, Chen, Levy, Lewis,
  Zettlemoyer, and Stoyanov}]{liu2019roberta}
Yinhan Liu, Myle Ott, Naman Goyal, Jingfei Du, Mandar Joshi, Danqi Chen, Omer
  Levy, Mike Lewis, Luke Zettlemoyer, and Veselin Stoyanov. 2019.
\newblock \href {http://arxiv.org/abs/1907.11692} {{RoBERTa}: A robustly
  optimized {BERT} pretraining approach}.
\newblock arXiv:1907.11692.

\bibitem[{Longpre et~al.(2019)Longpre, Lu, Tu, and
  DuBois}]{longpre2019exploration}
Shayne Longpre, Yi~Lu, Zhucheng Tu, and Chris DuBois. 2019.
\newblock An exploration of data augmentation and sampling techniques for
  domain-agnostic question answering.
\newblock \emph{arXiv preprint arXiv:1912.02145}.

\bibitem[{Maas et~al.(2011)Maas, Daly, Pham, Huang, Ng, and
  Potts}]{maas-etal-2011-learning}
Andrew~L. Maas, Raymond~E. Daly, Peter~T. Pham, Dan Huang, Andrew~Y. Ng, and
  Christopher Potts. 2011.
\newblock \href {https://aclanthology.org/P11-1015} {Learning word vectors for
  sentiment analysis}.
\newblock In \emph{Proceedings of the 49th Annual Meeting of the Association
  for Computational Linguistics: Human Language Technologies}, pages 142--150,
  Portland, Oregon, USA. Association for Computational Linguistics.

\bibitem[{Marelli et~al.(2014)Marelli, Menini, Baroni, Bentivogli, Bernardi,
  and Zamparelli}]{marelli-etal-2014-sick}
Marco Marelli, Stefano Menini, Marco Baroni, Luisa Bentivogli, Raffaella
  Bernardi, and Roberto Zamparelli. 2014.
\newblock \href
  {http://www.lrec-conf.org/proceedings/lrec2014/pdf/363_Paper.pdf} {A {SICK}
  cure for the evaluation of compositional distributional semantic models}.
\newblock In \emph{Proceedings of the Ninth International Conference on
  Language Resources and Evaluation ({LREC}'14)}, pages 216--223, Reykjavik,
  Iceland. European Language Resources Association (ELRA).

\bibitem[{McCoy et~al.(2019)McCoy, Pavlick, and Linzen}]{mccoy-etal-2019-right}
Tom McCoy, Ellie Pavlick, and Tal Linzen. 2019.
\newblock \href {https://doi.org/10.18653/v1/P19-1334} {Right for the wrong
  reasons: Diagnosing syntactic heuristics in natural language inference}.
\newblock In \emph{Proceedings of the 57th Annual Meeting of the Association
  for Computational Linguistics}, pages 3428--3448, Florence, Italy.
  Association for Computational Linguistics.

\bibitem[{McCreery et~al.(2020)McCreery, Katariya, Kannan, Chablani, and
  Amatriain}]{mccreery2020effective}
Clara~H. McCreery, Namit Katariya, Anitha Kannan, Manish Chablani, and Xavier
  Amatriain. 2020.
\newblock \href {https://doi.org/10.1145/3394486.3412861} {Effective transfer
  learning for identifying similar questions: Matching user questions to
  {COVID-19} {FAQs}}.
\newblock In \emph{Proceedings of the 26th ACM SIGKDD International Conference
  on Knowledge Discovery \& Data Mining}, pages 3458–--3465. Association for
  Computing Machinery.

\bibitem[{Miller et~al.(2020)Miller, Krauth, Recht, and
  Schmidt}]{miller2020effect}
John Miller, Karl Krauth, Benjamin Recht, and Ludwig Schmidt. 2020.
\newblock \href {https://proceedings.mlr.press/v119/miller20a.html} {The effect
  of natural distribution shift on question answering models}.
\newblock In \emph{Proceedings of the 37th International Conference on Machine
  Learning}, volume 119 of \emph{Proceedings of Machine Learning Research},
  pages 6905--6916. PMLR.

\bibitem[{Mosbach et~al.(2021)Mosbach, Andriushchenko, and
  Klakow}]{mosbach2020stability}
Marius Mosbach, Maksym Andriushchenko, and Dietrich Klakow. 2021.
\newblock \href {https://openreview.net/forum?id=nzpLWnVAyah} {On the stability
  of fine-tuning {BERT}: Misconceptions, explanations, and strong baselines}.
\newblock In \emph{International Conference on Learning Representations}.

\bibitem[{Ng et~al.(2020)Ng, Cho, and Ghassemi}]{ng-etal-2020-ssmba}
Nathan Ng, Kyunghyun Cho, and Marzyeh Ghassemi. 2020.
\newblock \href {https://doi.org/10.18653/v1/2020.emnlp-main.97} {{SSMBA}:
  Self-supervised manifold based data augmentation for improving out-of-domain
  robustness}.
\newblock In \emph{Proceedings of the 2020 Conference on Empirical Methods in
  Natural Language Processing (EMNLP)}, pages 1268--1283, Online. Association
  for Computational Linguistics.

\bibitem[{Nie et~al.(2020)Nie, Williams, Dinan, Bansal, Weston, and
  Kiela}]{nie-etal-2020-adversarial}
Yixin Nie, Adina Williams, Emily Dinan, Mohit Bansal, Jason Weston, and Douwe
  Kiela. 2020.
\newblock \href {https://doi.org/10.18653/v1/2020.acl-main.441} {Adversarial
  {NLI}: A new benchmark for natural language understanding}.
\newblock In \emph{Proceedings of the 58th Annual Meeting of the Association
  for Computational Linguistics}, pages 4885--4901, Online. Association for
  Computational Linguistics.

\bibitem[{Ott et~al.(2019)Ott, Edunov, Baevski, Fan, Gross, Ng, Grangier, and
  Auli}]{ott-etal-2019-fairseq}
Myle Ott, Sergey Edunov, Alexei Baevski, Angela Fan, Sam Gross, Nathan Ng,
  David Grangier, and Michael Auli. 2019.
\newblock \href {https://doi.org/10.18653/v1/N19-4009} {fairseq: A fast,
  extensible toolkit for sequence modeling}.
\newblock In \emph{Proceedings of the 2019 Conference of the North {A}merican
  Chapter of the Association for Computational Linguistics (Demonstrations)},
  pages 48--53, Minneapolis, Minnesota. Association for Computational
  Linguistics.

\bibitem[{Qu et~al.(2021)Qu, Shen, Shen, Sajeev, Chen, and Han}]{qu2021coda}
Yanru Qu, Dinghan Shen, Yelong Shen, Sandra Sajeev, Weizhu Chen, and Jiawei
  Han. 2021.
\newblock \href {https://openreview.net/forum?id=Ozk9MrX1hvA} {Co{DA}:
  Contrast-enhanced and diversity-promoting data augmentation for natural
  language understanding}.
\newblock In \emph{International Conference on Learning Representations}.

\bibitem[{Rajpurkar et~al.(2016)Rajpurkar, Zhang, Lopyrev, and
  Liang}]{rajpurkar-etal-2016-squad}
Pranav Rajpurkar, Jian Zhang, Konstantin Lopyrev, and Percy Liang. 2016.
\newblock \href {https://doi.org/10.18653/v1/D16-1264} {{SQ}u{AD}: 100,000+
  questions for machine comprehension of text}.
\newblock In \emph{Proceedings of the 2016 Conference on Empirical Methods in
  Natural Language Processing}, pages 2383--2392, Austin, Texas. Association
  for Computational Linguistics.

\bibitem[{Rashid et~al.(2021)Rashid, Lioutas, and
  Rezagholizadeh}]{rashid-etal-2021-mate}
Ahmad Rashid, Vasileios Lioutas, and Mehdi Rezagholizadeh. 2021.
\newblock \href {https://doi.org/10.18653/v1/2021.acl-long.86} {{MATE}-{KD}:
  Masked adversarial {TE}xt, a companion to knowledge distillation}.
\newblock In \emph{Proceedings of the 59th Annual Meeting of the Association
  for Computational Linguistics and the 11th International Joint Conference on
  Natural Language Processing (Volume 1: Long Papers)}, pages 1062--1071,
  Online. Association for Computational Linguistics.

\bibitem[{Rogers(2021)}]{rogers-2021-changing}
Anna Rogers. 2021.
\newblock \href {https://doi.org/10.18653/v1/2021.acl-long.170} {Changing the
  world by changing the data}.
\newblock In \emph{Proceedings of the 59th Annual Meeting of the Association
  for Computational Linguistics and the 11th International Joint Conference on
  Natural Language Processing (Volume 1: Long Papers)}, pages 2182--2194,
  Online. Association for Computational Linguistics.

\bibitem[{Sambasivan et~al.(2021)Sambasivan, Kapania, Highfill, Akrong,
  Paritosh, and Aroyo}]{sambasivan2021everyone}
Nithya Sambasivan, Shivani Kapania, Hannah Highfill, Diana Akrong, Praveen
  Paritosh, and Lora~M Aroyo. 2021.
\newblock \href {https://doi.org/10.1145/3411764.3445518} {“everyone wants to
  do the model work, not the data work”: Data cascades in high-stakes ai}.
\newblock In \emph{Proceedings of the 2021 CHI Conference on Human Factors in
  Computing Systems}, pages 1--15. Association for Computing Machinery.

\bibitem[{Sanh et~al.(2019)Sanh, Debut, Chaumond, and
  Wolf}]{sanh2019distilbert}
Victor Sanh, Lysandre Debut, Julien Chaumond, and Thomas Wolf. 2019.
\newblock \href {http://arxiv.org/abs/1910.01108} {{DistilBERT}, a distilled
  version of {BERT}: smaller, faster, cheaper and lighter}.
\newblock arXiv:1910.01108.

\bibitem[{Sennrich et~al.(2016)Sennrich, Haddow, and
  Birch}]{sennrich-etal-2016-improving}
Rico Sennrich, Barry Haddow, and Alexandra Birch. 2016.
\newblock \href {https://doi.org/10.18653/v1/P16-1009} {Improving neural
  machine translation models with monolingual data}.
\newblock In \emph{Proceedings of the 54th Annual Meeting of the Association
  for Computational Linguistics (Volume 1: Long Papers)}, pages 86--96, Berlin,
  Germany. Association for Computational Linguistics.

\bibitem[{Shakeri et~al.(2020)Shakeri, Nogueira~dos Santos, Zhu, Ng, Nan, Wang,
  Nallapati, and Xiang}]{shakeri-etal-2020-end}
Siamak Shakeri, Cicero Nogueira~dos Santos, Henghui Zhu, Patrick Ng, Feng Nan,
  Zhiguo Wang, Ramesh Nallapati, and Bing Xiang. 2020.
\newblock \href {https://doi.org/10.18653/v1/2020.emnlp-main.439} {End-to-end
  synthetic data generation for domain adaptation of question answering
  systems}.
\newblock In \emph{Proceedings of the 2020 Conference on Empirical Methods in
  Natural Language Processing (EMNLP)}, pages 5445--5460, Online. Association
  for Computational Linguistics.

\bibitem[{Shen et~al.(2020)Shen, Zheng, Shen, Qu, and Chen}]{shen2020simple}
Dinghan Shen, Mingzhi Zheng, Yelong Shen, Yanru Qu, and Weizhu Chen. 2020.
\newblock A simple but tough-to-beat data augmentation approach for natural
  language understanding and generation.
\newblock \emph{arXiv preprint arXiv:2009.13818}.

\bibitem[{Volpi et~al.(2018)Volpi, Namkoong, Sener, Duchi, Murino, and
  Savarese}]{volpi2018generalizing}
Riccardo Volpi, Hongseok Namkoong, Ozan Sener, John Duchi, Vittorio Murino, and
  Silvio Savarese. 2018.
\newblock \href
  {https://proceedings.neurips.cc/paper/2018/file/1d94108e907bb8311d8802b48fd54b4a-Paper.pdf}
  {Generalizing to unseen domains via adversarial data augmentation}.
\newblock \emph{Advances in neural information processing systems}, 31.

\bibitem[{Wang et~al.(2019)Wang, Singh, Michael, Hill, Levy, and
  Bowman}]{wang2018glue}
Alex Wang, Amanpreet Singh, Julian Michael, Felix Hill, Omer Levy, and
  Samuel~R. Bowman. 2019.
\newblock \href {https://openreview.net/forum?id=rJ4km2R5t7} {{GLUE}: A
  multi-task benchmark and analysis platform for natural language
  understanding}.
\newblock In \emph{International Conference on Learning Representations}.

\bibitem[{Wang et~al.(2020)Wang, Li, Wang, and Gong}]{wang2020neural}
Dongdong Wang, Yandong Li, Liqiang Wang, and Boqing Gong. 2020.
\newblock Neural networks are more productive teachers than human raters:
  Active mixup for data-efficient knowledge distillation from a blackbox model.
\newblock In \emph{Proceedings of the IEEE/CVF Conference on Computer Vision
  and Pattern Recognition}, pages 1498--1507.

\bibitem[{Wei and Zou(2019)}]{wei-zou-2019-eda}
Jason Wei and Kai Zou. 2019.
\newblock \href {https://doi.org/10.18653/v1/D19-1670} {{EDA}: Easy data
  augmentation techniques for boosting performance on text classification
  tasks}.
\newblock In \emph{Proceedings of the 2019 Conference on Empirical Methods in
  Natural Language Processing and the 9th International Joint Conference on
  Natural Language Processing (EMNLP-IJCNLP)}, pages 6382--6388, Hong Kong,
  China. Association for Computational Linguistics.

\bibitem[{Williams et~al.(2018)Williams, Nangia, and
  Bowman}]{williams-etal-2018-broad}
Adina Williams, Nikita Nangia, and Samuel Bowman. 2018.
\newblock \href {https://doi.org/10.18653/v1/N18-1101} {A broad-coverage
  challenge corpus for sentence understanding through inference}.
\newblock In \emph{Proceedings of the 2018 Conference of the North {A}merican
  Chapter of the Association for Computational Linguistics: Human Language
  Technologies, Volume 1 (Long Papers)}, pages 1112--1122, New Orleans,
  Louisiana. Association for Computational Linguistics.

\bibitem[{Wolf et~al.(2020)Wolf, Debut, Sanh, Chaumond, Delangue, Moi, Cistac,
  Rault, Louf, Funtowicz, Davison, Shleifer, von Platen, Ma, Jernite, Plu, Xu,
  Le~Scao, Gugger, Drame, Lhoest, and Rush}]{wolf-etal-2020-transformers}
Thomas Wolf, Lysandre Debut, Victor Sanh, Julien Chaumond, Clement Delangue,
  Anthony Moi, Pierric Cistac, Tim Rault, Remi Louf, Morgan Funtowicz, Joe
  Davison, Sam Shleifer, Patrick von Platen, Clara Ma, Yacine Jernite, Julien
  Plu, Canwen Xu, Teven Le~Scao, Sylvain Gugger, Mariama Drame, Quentin Lhoest,
  and Alexander Rush. 2020.
\newblock \href {https://doi.org/10.18653/v1/2020.emnlp-demos.6} {Transformers:
  State-of-the-art natural language processing}.
\newblock In \emph{Proceedings of the 2020 Conference on Empirical Methods in
  Natural Language Processing: System Demonstrations}, pages 38--45, Online.
  Association for Computational Linguistics.

\bibitem[{Wu et~al.(2019)Wu, Lv, Zang, Han, and Hu}]{wu2019conditional}
Xing Wu, Shangwen Lv, Liangjun Zang, Jizhong Han, and Songlin Hu. 2019.
\newblock Conditional {BERT} contextual augmentation.
\newblock In \emph{International Conference on Computational Science}, pages
  84--95. Springer International Publishing.

\bibitem[{Xie et~al.(2020)Xie, Dai, Hovy, Luong, and Le}]{xie2019unsupervised}
Qizhe Xie, Zihang Dai, Eduard Hovy, Thang Luong, and Quoc Le. 2020.
\newblock \href
  {https://proceedings.neurips.cc/paper/2020/file/44feb0096faa8326192570788b38c1d1-Paper.pdf}
  {Unsupervised data augmentation for consistency training}.
\newblock \emph{Advances in Neural Information Processing Systems},
  33:6256--6268.

\bibitem[{Yi et~al.(2021)Yi, Hou, Shang, Jiang, Liu, and
  Ma}]{yi2021reweighting}
Mingyang Yi, Lu~Hou, Lifeng Shang, Xin Jiang, Qun Liu, and Zhi-Ming Ma. 2021.
\newblock \href {https://openreview.net/forum?id=9G5MIc-goqB} {Reweighting
  augmented samples by minimizing the maximal expected loss}.
\newblock In \emph{International Conference on Learning Representations}.

\bibitem[{Zellers et~al.(2019)Zellers, Holtzman, Bisk, Farhadi, and
  Choi}]{zellers-etal-2019-hellaswag}
Rowan Zellers, Ari Holtzman, Yonatan Bisk, Ali Farhadi, and Yejin Choi. 2019.
\newblock \href {https://doi.org/10.18653/v1/P19-1472} {{H}ella{S}wag: Can a
  machine really finish your sentence?}
\newblock In \emph{Proceedings of the 57th Annual Meeting of the Association
  for Computational Linguistics}, pages 4791--4800, Florence, Italy.
  Association for Computational Linguistics.

\bibitem[{Zhang et~al.(2018)Zhang, Cisse, Dauphin, and
  Lopez-Paz}]{zhang2017mixup}
Hongyi Zhang, Moustapha Cisse, Yann~N Dauphin, and David Lopez-Paz. 2018.
\newblock \href {https://openreview.net/forum?id=r1Ddp1-Rb} {mixup: Beyond
  empirical risk minimization}.
\newblock In \emph{International Conference on Learning Representations}.

\bibitem[{Zhang et~al.(2015)Zhang, Zhao, and LeCun}]{zhang2015character}
Xiang Zhang, Junbo Zhao, and Yann LeCun. 2015.
\newblock \href
  {https://proceedings.neurips.cc/paper/2015/file/250cf8b51c773f3f8dc8b4be867a9a02-Paper.pdf}
  {Character-level convolutional networks for text classification}.
\newblock In \emph{Advances in Neural Information Processing Systems},
  volume~28, pages 649--657. Curran Associates, Inc.

\bibitem[{Zhang et~al.(2019)Zhang, Baldridge, and He}]{zhang-etal-2019-paws}
Yuan Zhang, Jason Baldridge, and Luheng He. 2019.
\newblock \href {https://doi.org/10.18653/v1/N19-1131} {{PAWS}: Paraphrase
  adversaries from word scrambling}.
\newblock In \emph{Proceedings of the 2019 Conference of the North {A}merican
  Chapter of the Association for Computational Linguistics: Human Language
  Technologies, Volume 1 (Long and Short Papers)}, pages 1298--1308,
  Minneapolis, Minnesota. Association for Computational Linguistics.

\end{thebibliography}
\bibliographystyle{acl_natbib}

\appendix
\clearpage

\section{Implementation Details}
\label{sec:supp_implement}

\subsection{Fine-tuning details}
We adopted the publicly available pre-trained {\Roberta} \cite{liu2019roberta} and {\Student} \cite{sanh2019distilbert}---using the Huggingface Transformers library \cite{wolf-etal-2020-transformers} and the Pytorch Lightning library\footnote{\url{https://github.com/PyTorchLightning/pytorch-lightning}}.

For the {\em test} settings, the model is evaluated on the development data once per epoch for small datasets and twice per epoch for large ones---i.e., SST-2, MNLI, QNLI, SQuAD, and HellaSwag. The best performing model is chosen for testing. Our learning rate schedule follows a
linear decay scheduler with a warm-up, specified as a ratio of the total number of training steps. Maximum number of epochs is set to 20 for all tasks  except SQuAD, following \cite{mosbach2020stability}. For large datasets, we early stop with a patience of 10. The learning rate, and the batch size are tuned for each task separately. The details of hyperparameters are summarized in Table~\ref{tab:glue_hyperparameters}. We ran {\RobertaB} experiments with the similar hyperparameters, but with these exceptions: On QNLI, learning rate, batch size, and weight decay are set to $\scism{3}{5}$, 64, and 0.1; warmup ratio is set to 0.06 on QQP.

For {\em dev} experiments, we follow CoDA \cite{qu2021coda} on the GLUE tasks. Specifically, we train the model for 5 epochs with a batch size of 32, learning rate $\scism{1}{5}$, warmup ratio 0.06, weight decay 0.1, and linear learning rate decay. For SQuAD, and HellaSwag, the hyperparameters are detailed in Table~\ref{tab:other_hyperparameters}.

All experiments were conducted on two Nvidia Tesla V100 GPUs.

\begin{table}[h!]
    \centering
    \begin{tabular}{l c c}
        \hline
        \small{{\bf Hyperparam.}} & {\bf SQuAD} & {\bf HellaSwag} \\
        \hline
        Learning rate & $\scism{1.5}{5}$ & $\scism{1.5}{5}$ \\
        Batch size & 16 & 32 \\
        Max length & 512 & 512 \\
        Max epochs & 3 & 20 \\
        Warmup ratio & 0.06 & 0.06 \\
        Grad. acc. steps & 4 & 1 \\
        Weight Decay & 0.01 & 0.01 \\
        temp. $\tau$ (for KD) & 5.0 & 10.0 \\
        \hline
    \end{tabular}
    \caption{Hyperparameters of {\Student} on two downstream tasks.}
    \label{tab:other_hyperparameters}
\end{table}

\begin{table*}[t!]
    \centering
    \begin{tabular}{l c c c c c c c c}
        \hline
        \small{{\bf Hyperparam.}} & \small{{\bf CoLA}} & \small{{\bf SST}} & \small{{\bf MRPC}} & \small{{\bf STS-B}} & \small{{\bf QQP}} & \small{{\bf MNLI-m/mm}} & \small{{\bf QNLI}} & \small{{\bf RTE}} \\
        \hline
        Learning rate & $\scism{1}{5}$ & $\scism{1}{5}$ & $\scism{1}{5}$ & $\scism{1}{5}$ & $\scism{1}{5}$ & $\scism{3}{5}$/$\scism{1}{5}$ & $\scism{5}{5}^*$ & $\scism{1}{5}$ \\
        Batch size & 32 & 64 & 16 & 32 & 64 & 64 & 128$^*$ & 32 \\
        Max length & 128 & 256 & 128 & 128 & 256 & 256 & 256 & 256 \\
        Warmup ratio & 0.1 & 0.06 & 0.06 & 0.06 & 0.1$^*$ & 0.08/0.06 & 0.08 & 0.06 \\
        Gradient acc. steps & 1 & 4 & 1 & 1 & 4 & 4 & 4 & 1 \\
        Weight Decay & 0.1 & 0.1 & 0.1 & 0.1 & 0.1 & 0.0/0.1 & 0.0$^*$ & 0.1 \\
        Softmax temp. $\tau$ (for KD) & 30.0 & 20.0 & 12.0$^*$ & 12.0 & 20.0 & 12.0 & 12.0 & 12.0 \\
        \hline
    \end{tabular}
    \caption{Hyperparameters of {\Student} on the GLUE benchmark. We used the same configuration for {\RobertaB} albeit with a few exceptions marked by ($^*$).}
    \label{tab:glue_hyperparameters}
\end{table*}

\subsection{Knowledge distillation details}
We implemented knowledge distillation by caching the teacher's logits prior to training.
We performed grid search to find the best softmax temperature $\tau$ from $\{$5.0, 10.0, 12.0, 20.0, 30.0$\}$. The value of $\tau$ used in our experiments are reported in \Cref{tab:other_hyperparameters,tab:glue_hyperparameters} for {\Student} and {\RobertaB}; with the exception $\tau=20.0$ on MRPC for {\RobertaB}. Loss weight $\alpha$, in Eq.~\eqref{eq:KD:task}, is set to 0.5 for all tasks except CoLA in which $\alpha=0.75$.

\section{OOD results}
\label{sec:ood_results}

\subsection{Distilled Mode} \label{ssec:ood_distilled}
OOD results for models trained in the distilled mode are presented in Table~\ref{tab:glue_ood_kd}.

\subsection{Standalone Mode} \label{ssec:ood_standalone}
Table~\ref{tab:glue_test_ood_self} presents OOD results for models trained using {\em test} settings, and Table~\ref{tab:glue_dev_ood_self} (complementary to Table~\ref{tab:glue_dev_base} in \S\ref{ssec:ood_glue}) presents OOD results for dev experiments.

\begin{table*}[t]
    \centering
    \begin{tabular}{l c c c c c c c c c}
        \hline
        \footnotesize{\emph{Trained On} $\rightarrow$} & \emph{\small{SST}} & \emph{\small{SST}} & \emph{\small{SST}} & \emph{\small{STS}} & \emph{\small{QQP}} & \emph{\small{QQP}} & \emph{\small{MNLI}} & \emph{\small{MNLI}} & \emph{\small{RTE}} \\
        \multirow{2}{*}{\small{{\bf Method}}} & \small{{\bf IMDb}} & \small{{\bf IMDb-Con.}} & \small{{\bf IMDb-CAD}} & \small{{\bf SICK}} & \small{{\bf MQP}} & \small{{\bf PAWS$_\text{QQP}$}} & \small{{\bf SciTail}} & \small{{\bf A-NLI}} & \small{{\bf HANS}} \\
         & \small{Acc} & \small{Acc} & \small{Acc} & \small{P/S} & \small{Acc/F$_1$} & \small{Acc} & \small{Acc} & \small{Acc} & \small{Acc} \\
        \hline
        {\small {\teacher}} & 93.7 & 92.0 & 94.0 & 84.3 & 71.6 & 43.6 & 82.0 & 45.9 & 81.8 \\
        \hline
        {\small {\student}} & 90.2 & 87.6 & 92.5 & 79.6 & 67.3 & 36.3 & 74.8 & 27.8 & {\bf 71.3} \\
        {\small KD} & 90.6 & 87.4 & 93.2 & 79.9 & 65.6 & 33.1 & 77.3 & 28.9 & 70.6 \\
        \hline
        \multicolumn{10}{c}{{\em EDA} \cite{wei-zou-2019-eda}} \\
        {\small Vanilla-DA} & 91.8 & 87.2 & 92.9 & 80.0 & 59.9 & {\bf 38.0} & 75.8 & 27.3 & 66.6 \\
        {\small {\Ours}} & 91.2 & 87.1 & {\bf 94.0} & 80.0 & 64.0 & 36.6 & 75.6 & 28.8 & 65.6 \\
        \hline
        \multicolumn{10}{c}{{\em Back-Translation}} \\
        {\small Vanilla-DA} & 92.2 & 87.9 & 92.1 & 80.3 & {\bf 69.6} & 35.0 & 76.5 & 27.9 & 68.0 \\
        {\small {\Ours}} & {\bf 92.4} & 87.9 & 92.8 & {\bf 81.2} & 68.7 & 35.2 & 77.6 & {\bf 30.4} & 70.5 \\
        \hline
        \multicolumn{10}{c}{{\em Masked-and-reconstruct}} \\
        {\small Vanilla-DA} & 91.8 & {\bf 88.8} & 92.9 & 80.4 & 68.5 & 33.7 & 77.4 & 28.5 & 69.3 \\
        {\small {\Ours}} & 92.0 & 88.0 & 92.5 & 80.7 & 68.8 & 35.3 & {\bf 78.2} & 29.9 & 70.9 \\
        \hline
    \end{tabular}
    \caption{OOD results of models whose in-domain test results are reported in \Cref{tab:glue_test} for the distilled mode. {\bf Bold} numbers indicate the best result across {\student} models.}
    \label{tab:glue_ood_kd}
\end{table*}

\begin{table*}[t]
    \centering
    \begin{tabular}{l c c c c c c c c c}
        \hline
        \footnotesize{\emph{Trained On} $\rightarrow$} & \emph{\small{SST}} & \emph{\small{SST}} & \emph{\small{SST}} & \emph{\small{STS}} & \emph{\small{QQP}} & \emph{\small{QQP}} & \emph{\small{MNLI}} & \emph{\small{MNLI}} & \emph{\small{RTE}} \\
        \multirow{2}{*}{\small{{\bf Method}}} & \small{{\bf IMDb}} & \small{{\bf IMDb-Con.}} & \small{{\bf IMDb-CAD}} & \small{{\bf SICK}} & \small{{\bf MQP}} & \small{{\bf PAWS$_\text{QQP}$}} & \small{{\bf SciTail}} & \small{{\bf A-NLI}} & \small{{\bf HANS}} \\
         & \small{Acc} & \small{Acc} & \small{Acc} & \small{P/S} & \small{Acc/F$_1$} & \small{Acc} & \small{Acc} & \small{Acc} & \small{Acc} \\
        \hline
        {\small {\RobB}} & 92.2 & 89.1 & 94.3 & 80.6 & 70.7 & 38.6 & 78.5 & 31.4 & 78.5 \\
        \hline
        {\small Self-KD} & \textbf{92.6} & 89.1 & {\bf 95.0} & 80.2 & 70.9 & 37.6 & {\bf 79.4} & 32.1 & 79.5 \\
        {\small \, + Vanilla-DA} & 91.8 & 88.8 & 94.8 & 81.5 & 71.4 & 38.8 & 78.4 & 31.5 & 79.3 \\
        {\small \, + {\Ours}} & 92.0 & {\bf 89.6} & 94.8 & {\bf 81.7} & {\bf 72.1} & {\bf 39.4} & 79.1 & {\bf 32.7 } & 80.1 \\
        \hline
        {\small CT + Vanilla-DA} & 90.6 & 88.1 & 92.1 & 76.6 & 70.6 & 38.3 & 76.6 & 30.3 & 78.4 \\
        {\small CT + {\Ours}} & 92.2 & 88.6 & 93.7 & 79.4 & 70.7 & 38.8 & 77.0 & 31.6 & {\bf 80.2} \\
        \hline
    \end{tabular}
    \caption{OOD results of models whose in-domain test results are reported in Table~\ref{tab:glue_test_base} for the standalone experiment. {\bf Bold} numbers indicate the best result.}
    \label{tab:glue_test_ood_self}
\end{table*}

\begin{table*}[t]
    \centering
    \begin{tabular}{l c c c c c c c}
        \hline
        \footnotesize{\emph{Trained On} $\rightarrow$} & \emph{\small{SST}} & \emph{\small{SST}} & \emph{\small{SST}} & \emph{\small{MNLI}} & \emph{\small{MNLI}} & \emph{\small{MNLI}} & \emph{\small{RTE}} \\
        \multirow{2}{*}{\small{{\bf Method}}} & \small{{\bf IMDb}} & \small{{\bf IMDb-Con.}} & \small{{\bf IMDb-CAD}} & \small{{\bf SciTail}} & \small{{\bf A-NLI}} & \small{{\bf HANS}} & \small{{\bf HANS}} \\
         & \small{Acc} & \small{Acc} & \small{Acc} & \small{Acc} & \small{Acc} & \small{Acc} \\
        \hline
        {\small {\RobB}} & {\bf 91.9 {\scriptsize $\pm$ 0.3}} & 90.0 {\scriptsize $\pm$ 0.4} & 94.1 {\scriptsize $\pm$ 0.4} & {\bf 80.1 {\scriptsize $\pm$ 0.4}} & 31.0 {\scriptsize $\pm$ 0.6} & 73.7 {\scriptsize $\pm$ 0.7} & 78.3 {\scriptsize $\pm$ 0.4}  \\
        \hline
        {\small HiddenCut}$^{\scriptscriptstyle \spadesuit}$ & - & 87.8 & 90.4 & - & {\bf 32.8} & 71.2$^*$ & - \\
        {\small MMEL}$^{\scriptscriptstyle \dagger}$ & 91.6 {\scriptsize $\pm$ 0.1} & 90.5 {\scriptsize $\pm$ 0.7} & 94.5 {\scriptsize $\pm$ 0.4} & 79.7 {\scriptsize $\pm$ 0.3} & 31.4 {\scriptsize $\pm$ 0.6} & 74.5 {\scriptsize $\pm$ 0.6} & 78.3 {\scriptsize $\pm$ 0.3} \\
        \hline
        {\small Self-KD} & {\bf 91.9 {\scriptsize $\pm$ 0.3}} & 90.3 {\scriptsize $\pm$ 0.5} & 94.4 {\scriptsize $\pm$ 0.4} & 79.9 {\scriptsize $\pm$ 0.3} & 30.9 {\scriptsize $\pm$ 0.4} & 73.5 {\scriptsize $\pm$ 0.7} & 78.2 {\scriptsize $\pm$ 0.4} \\
        {\small \, + Vanilla-DA} & 91.6 {\scriptsize $\pm$ 0.4} & 90.2 {\scriptsize $\pm$ 0.4} & 94.3 {\scriptsize $\pm$ 0.3} & 79.3 {\scriptsize $\pm$ 0.4} & 31.3 {\scriptsize $\pm$ 0.5} & 73.9 {\scriptsize $\pm$ 0.4} & 77.8 {\scriptsize $\pm$ 0.3} \\
        {\small \, + {\Ours}} & 91.7{\scriptsize $\pm$ 0.2} & {\bf 90.6{\scriptsize $\pm$ 0.2}} & {\bf 94.8{\scriptsize $\pm$ 0.2}} & 79.4 {\scriptsize $\pm$ 0.1} & 31.8 {\scriptsize $\pm$ 0.4} & {\bf 74.6 {\scriptsize $\pm$ 0.3}} & {\bf 78.4 {\scriptsize $\pm$ 0.2}} \\
        \hline
    \end{tabular}
    \caption{OOD results of models with {\em dev} settings in the standalone mode, same models whose results are reported in Table~\ref{tab:glue_dev_base}. ($^{\scriptscriptstyle \spadesuit}$) denotes results are taken verbatim from: HiddenCut \cite{chen-etal-2021-hiddencut}. ($^{\scriptscriptstyle \dagger}$) indicates the results are obtained from our implementation of MMEL \cite{yi2021reweighting}. {\bf Bold} numbers indicate the best result.}
    \label{tab:glue_dev_ood_self}
\end{table*}

\end{document}